\documentclass[conference]{IEEEtran}

\usepackage{cite}
\usepackage{algorithmic}
\usepackage{graphicx}
\usepackage{textcomp}
\usepackage{xcolor}
\usepackage{url,hyperref}
\usepackage{booktabs,multirow}
\usepackage{amsmath,amssymb,amsfonts}
\usepackage{graphicx,caption,subcaption}

\DeclareRobustCommand{\IEEEauthorrefmark}[1]{\smash{\textsuperscript{\footnotesize #1}}}

\begin{document}

\title{Higher-order Neural Additive Models: An Interpretable Machine Learning Model with Feature Interactions}

\author{
    \IEEEauthorblockN{Minkyu Kim\IEEEauthorrefmark{1}, Hyun-Soo Choi\IEEEauthorrefmark{1}\IEEEauthorrefmark{2}\IEEEauthorrefmark{*} and Jinho Kim\IEEEauthorrefmark{3}\IEEEauthorrefmark{*}}
    \IEEEauthorblockA{\IEEEauthorrefmark{1} \textit{Ziovision Co., Ltd.}, \textit{Republic of Korea}}
    \IEEEauthorblockA{\IEEEauthorrefmark{2} \textit{Seoul National University of Science and Technology}, \textit{Republic of Korea}}
    \IEEEauthorblockA{\IEEEauthorrefmark{3} \textit{Kangwon National University}, \textit{Republic of Korea}}
    \IEEEauthorblockA{\IEEEauthorrefmark{*} Corresponding author}
    \IEEEauthorblockA{
        \href{mailto:minkyu.kim@ziovision.co.kr}{minkyu.kim@ziovision.co.kr},
        \href{mailto:choi.hyunsoo@seoultech.ac.kr}{choi.hyunsoo@seoultech.ac.kr},
        \href{mailto:jhkim@kangwon.ac.kr}{jhkim@kangwon.ac.kr}
    }
}

\maketitle

\begin{abstract}
Neural Additive Models (NAMs) have recently demonstrated promising predictive performance while maintaining interpretability. However, their capacity is limited to capturing only first-order feature interactions, which restricts their effectiveness on real-world datasets. To address this limitation, we propose Higher-order Neural Additive Models (HONAMs), an interpretable machine learning model that effectively and efficiently captures feature interactions of arbitrary orders. HONAMs improve predictive accuracy without compromising interpretability, an essential requirement in high-stakes applications. This advantage of HONAM can help analyze and extract high-order interactions present in datasets. The source code for HONAM is publicly available at \url{https://github.com/gim4855744/HONAM/}.
\end{abstract}


\section{Introduction}

Black-box models, such as deep neural networks, have demonstrated superior predictive performance across diverse fields, including computer vision, natural language processing, and recommender systems. However, their decision-making processes are inherently opaque. Recently, various explainable artificial intelligence (XAI) methods have been developed to uncover these processes by identifying critical features or regions influencing predictions. Nevertheless, applications of XAI in high-stakes domains, such as healthcare and social safety, remain limited because these methods frequently provide inaccurate or unfaithful explanations of the underlying models’ behaviors \cite{rudin2018please,rudin2019stop}. Note that, in this paper, we clearly differentiate between explanation methods (e.g., post-hoc feature attribution methods like SHAP and LIME) and interpretable models (e.g., glass-box models like linear models and generalized additive models).

Recently, Neural Additive Models (NAMs) have been introduced to enhance Generalized Additive Models (GAMs) by integrating neural networks \cite{agarwal2020neural}. NAM consists of a linear combination of neural networks, each associated with an individual input feature. Although NAM offers competitive performance compared to extreme gradient boosting (XGBoost) and multi-layer perceptrons (MLPs) along with interpretability, they have a notable limitation: they can only capture first-order feature interactions. Specifically, a prediction of NAM can be decomposed into additive contributions from individual features. However, real-world datasets frequently involve higher-order interactions—effects stemming from combinations of multiple features—that NAM cannot capture. This limitation leads to suboptimal predictive performance and lower-quality interpretations.

To address this limitation, we introduce a novel interpretable machine learning model called higher-order neural additive models (HONAMs). Since NAM is unsuitable for capturing high-order feature interactions, we restructure the additive framework of NAM to effectively capture interactions of arbitrary order. Additionally, we propose a new feature interaction method designed to address interpretability challenges and computational costs associated with existing methods. HONAM consists of a linear combination of neural networks each corresponding to individual input and the proposed interaction method models feature interactions of arbitrary-order. Therefore, HONAM can capture non-linear high-order feature interactions while being interpretable.

We conduct extensive experiments using various real-world datasets to evaluate the effectiveness of HONAM. The experimental results demonstrate that HONAM outperforms existing interpretable models and achieves competitive performance compared to black-box models. By visualizing HONAM’s predictions, particularly first- and second-order feature interactions, we show that HONAM effectively identifies valuable patterns in second-order interactions that NAM cannot capture. This emphasizes HONAM’s suitability for high-stakes domains requiring both strong predictive performance and high-quality interpretations. In addition, this advantage of HONAM can be beneficial in data mining tasks such as bias detection \cite{agarwal2020neural,tan_distill-and-compare_2018} and scientific discovery \cite{pedersen_hierarchical_2019,hastie_generalized_1995}, where interpretable models have already been successfully applied.

\section{Related Works}

    \subsection{Feature Interaction Methods}

    An $n$-order feature interaction reflects how combination of $n$ features influence the model output. For example, consider three features: $x_1$, $x_2$, and $x_3$. First-order interactions indicate the individual effects of each feature on the output. Second-order interactions represent the combined effects of two distinct features ($x_1 \times x_2$, $x_1 \times x_3$, $x_2 \times x_3$) on the output. Similarly, the third-order interaction describe the joint effect of all three features ($x_1 \times x_2 \times x_3$) on the output.

    Machine learning methods for explicitly capturing feature interactions have been widely studied. Factorization Machine (FM) \cite{rendle2010factorization} simultaneously capture first- and second-order feature interactions, demonstrating strong performance, particularly in recommender systems. Higher-order FM (HOFM) \cite{blondel2016higer} extend FM to capture interactions beyond the second order. Attentional FM (AFM) \cite{xiao2017attentional} integrate attention mechanisms into FM to weigh feature interactions dynamically. Recently, neural networks have been utilized to capture higher-order interactions \cite{cheng2016wide, guo2017deepfm}. Cross Network (CrossNet) \cite{wang2017deep} employs a multi-layer structure similar to an MLP but multiplies the sum of first-order features in every layer without activation functions, allowing a $t$-layer CrossNet to capture interactions up to the $t^{\text{th}}$ order. Several approaches have also been developed to simultaneously capture feature interactions across multiple orders \cite{lian2018xdeepfm, kim2020combining}. Additionally, Adaptive Factorization Network (AFN) \cite{cheng2020adaptive} adaptively selects interaction order during the learning process.

    Feature interaction methods have demonstrated success in predictive tasks such as recommendation and regression. Despite their effectiveness, these methods face several challenges. Many existing methods rely on linear interactions, limiting their ability to model complex, nonlinear relationships, thus restricting their expressive power. While some approaches utilize deep neural networks to capture higher-order interactions, they typically lack interpretability. Additionally, most previous research on feature interactions has primarily focused on predictive performance, often neglecting the interpretability inherent in linear models.

    \subsection{Generalized Additive Models}

    GAM is a leading framework for inherently interpretable (i.e., transparent) model \cite{hastie1987generalized, lou2012intelligible, chang2021interpretable}. GAM makes outputs as a linear combination of univariate functions, each reflecting the contribution of a single feature. GAM is particularly suitable for high-stakes domains due to their interpretability and strong predictive performance \cite{chang2021interpretable, caruana2015intelligible}. GA$^2$M \cite{lou2013accurate} extends GAM by incorporating second-order (pairwise) feature interactions. Explainable Boosting Machine (EBM) \cite{nori2019interpret}, a tree-based GAM, surpasses traditional GAMs and achieves competitive accuracy compared to tree-based ensemble models such as random forests and XGBoost. However, extending tree-based models to multi-task, multi-label, or transfer learning is challenging \cite{agarwal2020neural, chang2022node}. Generalized Additive Neural Network (GANN) \cite{potts1999generalized} utilizes shallow neural networks to construct nonlinear GAM, whereas the recently introduced Neural Additive Model (NAM) \cite{agarwal2020neural} leverages deep neural networks, capturing more complex nonlinear relationships. Although NAM outperforms other GAMs, it is limited to capturing only first-order feature interactions. NodeGAM and NodeGA$^2$M \cite{chang2022node} are neural tree-based GAMs; however, they are restricted to first-order and second-order interactions, respectively. Furthermore, existing GAMs rely on manually designed features to represent high-order interactions, which demands domain expertise and is time-consuming. Therefore, they usually employ all combinatorial features as input, but this leads to exponential increases in both model size and computation time as the interaction order increases. In contrast, our proposed HONAM maintains a consistent model size and scales linearly in computational complexity with the number of features and interaction order through an efficient feature interaction module.

    \subsection{Explainable AI}

    Deep neural networks have demonstrated superior performance in various fields, yet their decision-making processes remain difficult to understand. To address this problem, numerous XAI methods have been proposed. Several studies employ attention mechanisms to evaluate feature importance or select salient features \cite{gui2019afs, vskrlj2020feature}. For example, TabNet \cite{arik2021tabnet} uses a soft mask, analogous to attention scores, to identify important features. While attention mechanisms effectively highlight key features, their explanations may not always reliably reflect the model’s true prediction processes \cite{serrano2019attention,grimsley2020attention,tutek2020staying}.

    Recent XAI methods follow the post-hoc model-agnostic manner, applicable to any machine learning model without affecting performance. For instance, Layer-wise Relevance Propagation (LRP) \cite{bach2015pixel} decomposes model outputs into relevance scores, propagating them back to the input layer to indicate feature importance. However, LRP can produce misleading explanations. To address this, Deep Learning Important FeaTures (DeepLIFT) \cite{shrikumar2017learning} employs a reference-based strategy. Local Interpretable Model-agnostic Explanations (LIME) \cite{ribeiro2016why} approximates the predictions of a black-box model locally using an interpretable surrogate model, effectively explaining individual predictions when the approximation is accurate. SHapley Additive Explanations (SHAP) \cite{lundberg2017unified}, a game-theoretic method, assesses feature influence by measuring prediction changes resulting from feature omission. Despite these advancements, XAI methods can still yield explanations that do not reliably reflect true model behavior \cite{rudin2018please,rudin2019stop,akhavan_evaluating_2025,rahnama_study_2019,montavon_methods_2018,ghorbani_interpretation_2019,rahnama_can_2024,liu_synthetic_2021,akhavan_rahnama_blame_2023}, thereby limiting their application in high-stakes domains.

    In recent years, counterfactual (CF) example methods—generating a data point minimally modified from an original data point to yield a different prediction—have gained considerable attention. \cite{tolomei2017interpretable} introduced CF methods tailored for tree-based models, while \cite{mothilal2020explaining} developed a method to generate actionable and diverse CF examples specifically for differentiable models, such as neural networks. Additional CF methods for neural networks have also been proposed \cite{karimi2020model,lucic2022focus}, including reinforcement-learning-based methods by \cite{chen2022relax}. CF methods are advantageous over traditional feature attribution approaches like SHAP due to their inherent fidelity to the prediction model. Nevertheless, CF methods face several limitations: (1) difficulty in accurately determining feature importance, (2) the derived feature importance may not genuinely reflect the model’s decision-making process, and (3) the true contribution of individual features remains unclear.

\section{Higher-order Neural Additive Models}

    \subsection{Problem Statements on Neural Additive Models}

    In this paper, we address limitations of NAM. The original NAM formulation is defined as follows:
    \begin{equation}
    \hat{y} = \sum_{i=1}^{m} f_{i} \left( x_{i} \right) + b,
    \label{eq:original_old}
\end{equation}\noindent
    where $x_i \in \mathbb{R}^{m}$ represents the $i^\text{th}$ input feature, $f_{i}: \mathbb{R} \rightarrow \mathbb{R}$ is the MLP corresponding to the $i^\text{th}$ feature, $b \in \mathbb{R}$ is the output bias, and $m$ denotes the number of features. As shown in \eqref{eq:original_old}, NAM linearly combines individual feature-specific MLPs, where each MLP output directly reflects the contribution of its feature. However, simply summing all $f_i \left(x_i \right)$ restricts NAM to capturing only first-order interactions, resulting in limited predictive performance and interpretability. To address this limitation, we propose HONAM, a method capable of capturing feature interactions of arbitrary order.

    \subsection{Transformation of Neural Additive Models}

    Our goal is to enable NAM to model higher-order feature interactions; however, the original NAM structure is unsuitable for this purpose, as it outputs scalar values for each feature-specific MLP. Effective modeling of feature interactions requires vector outputs. Therefore, we introduce a modified NAM structure better suited to capturing feature interactions, defined as follows:
    \begin{equation}
    \mathcal{F} (\textbf{x}) = \overset{m}{\underset{i=1}{\Arrowvert}} f_{i} \left( x_{i} \right),
    \label{eq:new_nam}
\end{equation}\noindent
    where $\Arrowvert$ represents the stacking operator for row vectors, $\mathbf{x} \in \mathbb{R}^m$ denotes the input features, and $f_{i}: \mathbb{R} \rightarrow \mathbb{R}^{k}$ is the MLP corresponding to the $i^{\text{th}}$ feature. Unlike the original NAM, our modified NAM produces a $k$-dimensional vector output for each feature. Consequently, $\mathcal{F}: \mathbb{R}^{m} \rightarrow \mathbb{R}^{m \times k}$ denotes the function that stacks these representation vectors into a matrix.

    \subsection{Modeling High-order Feature Interactions}

    \begin{figure}[!t]
    \centering
    \includegraphics[width=.9\linewidth]{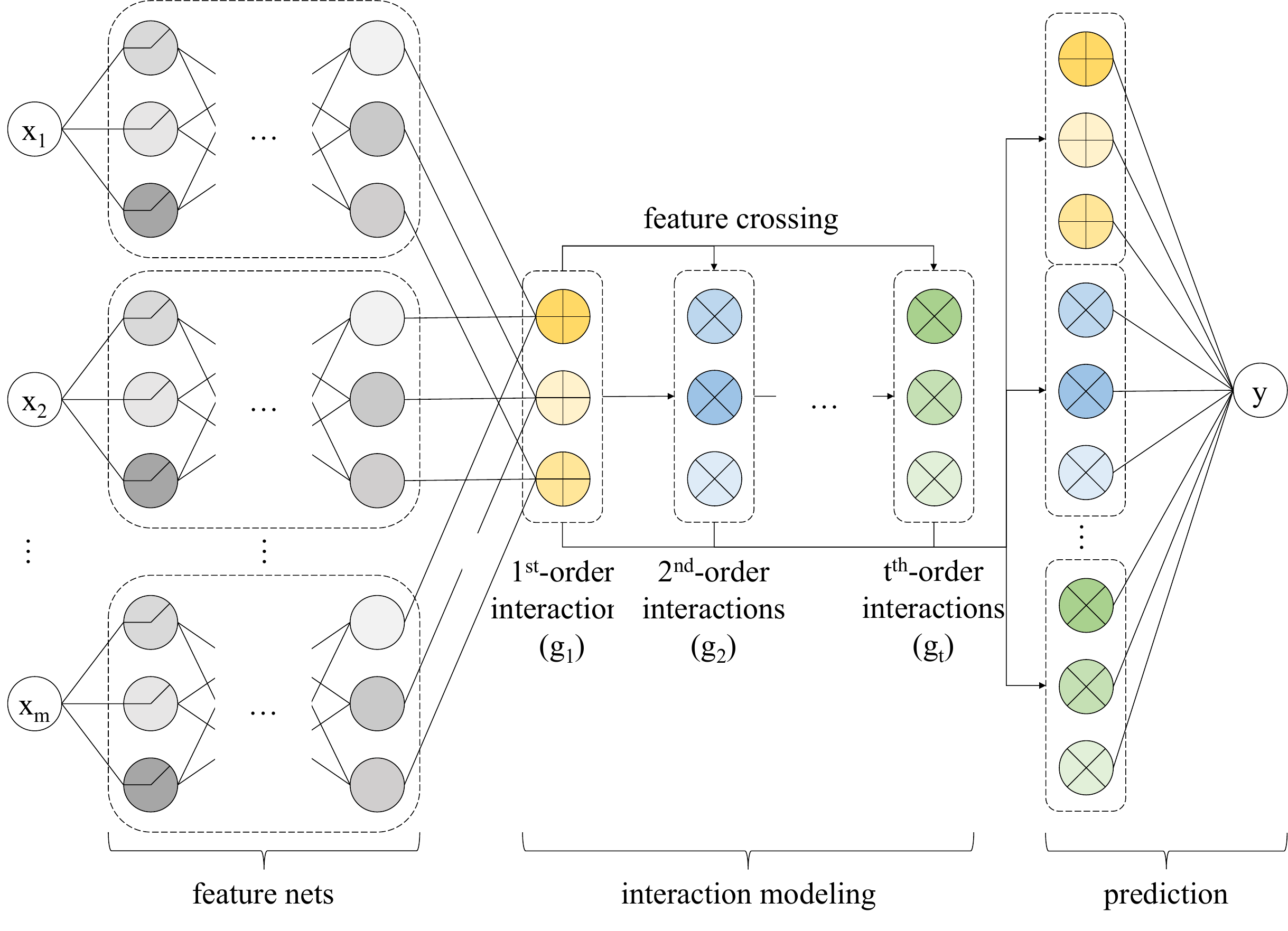}
    \caption{Architecture of HONAM. Different colors indicate different interaction orders.}
    \label{fig:honam}
\end{figure}
    \begin{figure}[!t]
    \centering
    \begin{minipage}{0.35\linewidth}
        \begin{subfigure}{\linewidth}
            \centering
            \includegraphics[width=\textwidth]{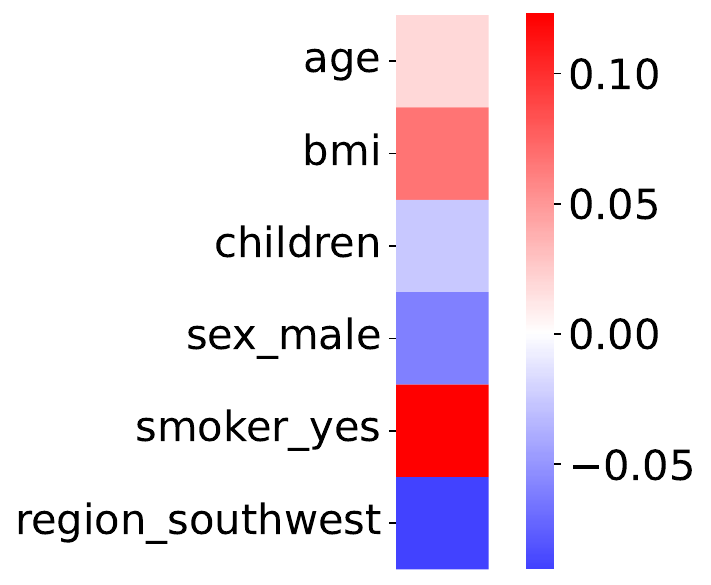}
            \caption{1$^{\text{st}}$-order features}
            \label{fig:insurance_first}
        \end{subfigure}
    \end{minipage}
    \begin{minipage}{0.55\linewidth}
        \vspace{1cm}
        \begin{subfigure}{\linewidth}
            \centering
            \includegraphics[width=\textwidth]{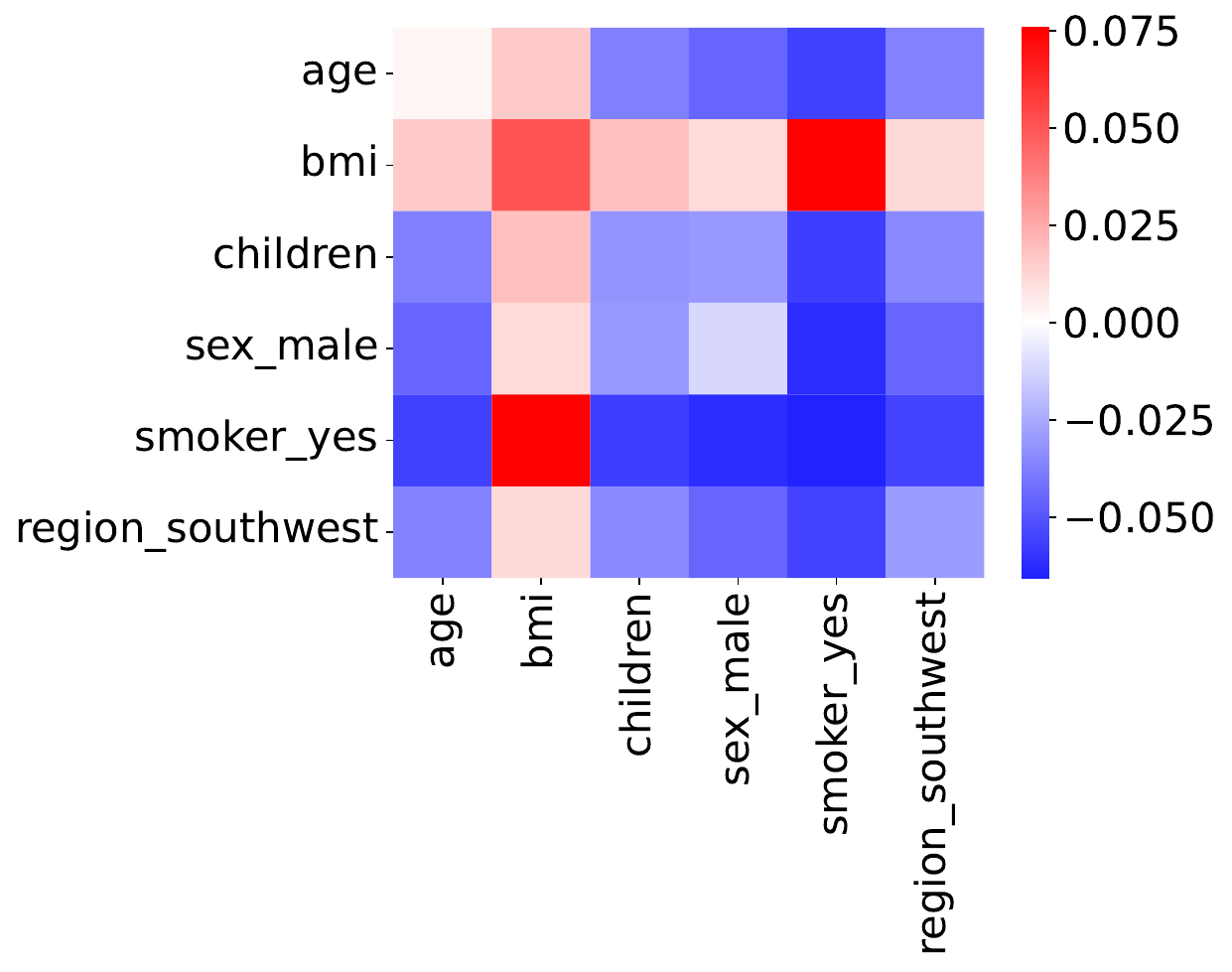}
            \caption{2$^{\text{nd}}$-order features}
            \label{fig:insurance_second}
        \end{subfigure}
    \end{minipage}
    \caption{Feature contributions on the Insurance dataset learned by CrossNet. Red and blue cells indicate features that have positive and negative effects, respectively.}
    \label{fig:interpretation_insurance}
\end{figure}

    Our objective in developing HONAM is to extend NAM capable of capturing high-order feature interactions. To achieve this, we propose an architecture consisting of NAM cascaded with a high-order feature interaction module, as illustrated in Fig.~\ref{fig:honam}. The resulting HONAM is defined as follows:
    \begin{gather}
    \hat{y} = \left( \overset{t}{\underset{i=1}{\arrowvert}} g_{i} \left( \mathbf{Z} \right) \right) \mathbf{W}^{(out)} + \mathbf{b}^{(out)},\label{eq:pred}\\
    \mathbf{Z}=\mathcal{F} \left( \mathbf{x} \right),
\end{gather}\noindent
    where $\arrowvert$ denotes the concatenation operator, $g_i \left( \cdot \right)$ denotes the feature interaction module responsible for modeling $i^\text{th}$-order interactions, $t$ indicates the maximum order of feature interactions considered, and $\mathbf{Z}$ represents the output of modified NAM defined in (\ref{eq:new_nam}). Additionally, $\mathbf{W}^{(out)} \in \mathbb{R}^{tk \times o}$ and $\textbf{b}^{(out)} \in \mathbb{R}^o$ represent the output weight and bias, respectively, where $o$ is the dimension of the output.

    The high-order feature interaction module in \eqref{eq:pred} can be implemented using CrossNet \cite{wang2017deep}, defined as follows:
    \begin{eqnarray}
    g_{i} \left( \mathbf{Z} \right) &=& \left( g_1 \left( \mathbf{Z} \right) \odot g_{i-1} \left( \mathbf{Z} \right) \right) \mathbf{W}_i, \\
    g_{1} \left( \mathbf{Z} \right) &=& \sum_{j=1}^{m} \mathbf{z}_j,\\
    g_{0} \left( \mathbf{Z} \right) &=& \mathbf{1},
\end{eqnarray}\noindent
    where $\odot$ denotes the Hadamard product, and $\mathbf{W}_i \in \mathbb{R}^{k \times k}$ represents the trainable weight for the $i^\text{th}$ layer. Although CrossNet effectively captures high-order feature interactions, it faces interpretability challenges. Specifically, CrossNet includes powered terms of features, such as $x_1^2$, $x_2^4$, or $x_1 x_2^3$, complicating the interpretation of predictions. For instance, Fig.~\ref{fig:insurance_first} and Fig.~\ref{fig:insurance_second} illustrate heat maps of contribution values for first-order and second-order features in the Insurance dataset \cite{insurance} learned by CrossNet. Interpreting squared features like $(sex=male)^2$ or $(smoker=yes)^2$ is difficult and lacks intuitive meaning. In addition, whereas a first-order feature $(smoker=yes)$ has a strongly positive contribution, its corresponding squared feature, $(smoker=yes)^2$, exhibit a negative contribution. As shown in Fig.~\ref{fig:interpretation_insurance}, these powered terms can generate conflicting contribution values compared to their original first-order counterparts, thereby complicating the understanding of actual feature effects.

    To overcome the aforementioned problem, we adopt a straightforward approach to capture high-order feature interactions. Specifically, we enumerate all possible combinations of distinct features, thus avoiding the generation of powered terms, following the approach suggested in \cite{rendle2010factorization}. Formally, this approach is defined as follows:
    \begin{equation}
    g_t \left ( \mathbf{Z} \right ) = \sum_{j_{1}=1}^{m} \sum_{j_{2}=j_{1}+1}^{m} \ldots \sum_{j_{t}=j_{t-1}+1}^{m} \bigodot_{l=1}^{t} \mathbf{z}_{j_{l}},
    \label{eq:simple_hofm}
\end{equation}\noindent
    where $t$ represents the order of feature interactions. This method enumerates all possible combinations of $t^\text{th}$-order features. However, simply enumerating these interactions has exponential time complexity $O(km^t)$, where $k$ is the dimension of feature representation vectors. Consequently, this approach results in slow training and inference times.

    \textbf{Proposition~1}. To alleviate the computational complexity of the simple enumerating method in \eqref{eq:simple_hofm}, we propose a recursive formulation for efficiently computing high-order feature interactions, defined as follows:
    \begin{eqnarray}
    g_{t} \left( \textbf{Z} \right) &=& \frac{1}{t} \sum_{i=1}^{t} \left(-1\right)^{i+1} g_{1} \left( \textbf{Z}^{i} \right) \circ g_{t-i} \left( \textbf{Z} \right), \label{eq:recursive_hofm} \\
    g_{1} \left( \textbf{Z} \right) &=& \sum_{j=1}^{m} \textbf{z}_j,\\
    g_{0} \left( \textbf{Z} \right) &=& \textbf{1}.
\end{eqnarray}\noindent
    Then, the recursive formulation in \eqref{eq:recursive_hofm} is equivalent to the simple enumerating method described in \eqref{eq:simple_hofm}. The proof of \textbf{Proposition~1} is provided in Appendix~\ref{sec:proof1}.

    Using dynamic programming, \eqref{eq:recursive_hofm} has a time complexity of $O(kmt)$. Thus, we can efficiently compute $t^\text{th}$-order feature interactions without generating powered terms in linear time.

    \subsection{Interpretability of HONAM}

    Our HONAM provides superiority compared to existing interpretable models for the following reasons: (1) HONAM captures complex nonlinear patterns by leveraging neural networks. (2) HONAM effectively captures feature interactions of arbitrary-orders through our proposed feature interaction module.

    Traditional interpretable models are primarily linear or tree-based, and recent approaches such as NAM, NodeGAM, and NodeGA$^2$M are limited to capturing only first- or second-order interactions. To the best of our knowledge, HONAM is the first interpretable model capable of capturing high-order feature interactions in an end-to-end manner.

    We can directly obtain the contributions of input features through forward propagation of HONAM. We denote the representation vector of feature $x_i$, computed by its corresponding MLP, as $\mathbf{z}_i$. The contribution of a first-order feature $x_i$ is computed as $\mathbf{z}_i \cdot \mathbf{w}^{(out)}_{0:k}$. Similarly, the contribution of a second-order feature $(x_i \times x_j)$ is computed as $(\mathbf{z}_i \odot \mathbf{z}_j) \cdot \mathbf{w}^{(out)}_{k:2k}$. Unlike NodeGAM, HONAM does not require purification for interpretability due to three reasons: (1) each feature has a single unique representation vector, (2) this representation is directly connected to the output layer, and (3) our feature interaction module captures only single-term interactions between distinct features (e.g., $x_1 \times x_2$ or $x_1 \times x_2 \times x_3$) rather than multi-term polynomials (e.g., $x_1x_2 + x_2$ or $x_2x_3 + x_2$).

    \subsection{Relationship with Higher-order Factorization Machines}

    HOFM can be viewed as a special case of HONAM. Specifically, if each feature network is linear and the output weight is an all-one matrix, HONAM becomes equivalent to HOFM. Additionally, we provide theoretical proof demonstrating that our recursive formulation is functionally equivalent to the enumeration method in Appendix~\ref{sec:proof1}.

\section{Experiments}

    \subsection{Datasets}\label{sec:dataset}

    \begin{table}[!t]
    \centering
    \resizebox{\linewidth}{!}{
        \begin{tabular}{lrrrc}
            \toprule
            & \# of samples & \# of features & positive rate & task \\
            \midrule
            California Housing &	 20,640 & 8 & - & regression \\
            Insurance & 1,338 & 6 & - & regression \\
            House Prices & 1,460 & 60 & - & regression \\
            Bikeshare & 17,389 & 16 & - & regression \\
            Year & 515,345 & 90 & - & regression \\
            FICO & 10,459 & 23 & 0.478 & classification \\
            Credit & 284,807 & 29 & 0.002 & classification \\
            SUPPORT2 & 9,105 & 29 & 0.259 & classification \\
            MIMIC-III & 27,348 & 57 & 0.098 & classification \\
            Click & 1M & 11 & 0.500 & classification \\
            \bottomrule
        \end{tabular}
    }
    \caption{Dataset statistics.}
    \label{tab:dataset}
\end{table}
    \begin{table*}[!t]

    \centering
    
    \resizebox{.75\linewidth}{!}{
        \begin{tabular}{lcccccccccc}
            \toprule
            & \multicolumn{2}{c}{California Housing}
            & \multicolumn{2}{c}{Insurance}
            & \multicolumn{2}{c}{House Prices}
            & \multicolumn{2}{c}{Bikeshare}
            & \multicolumn{2}{c}{Year} \\
            & R-squared & R-absolute
            & R-squared & R-absolute
            & R-squared & R-absolute
            & R-squared & R-absolute
            & R-squared & R-absolute\\
            \midrule
            \multirow{2}{*}{XGBoost}
            & \textbf{0.825} & \textbf{0.649}
            & 0.860 & \textbf{0.735}
            & \textbf{0.900} & \textbf{0.711}
            & \textbf{0.948} & \textbf{0.816}
            & 0.297 & 0.201\\
            & ($\pm$0.001) & ($\pm$0.002)
            & ($\pm$0.009) & ($\pm$0.012)
            & ($\pm$0.016) & ($\pm$0.011)
            & ($\pm$0.003) & ($\pm$0.002)
            & ($\pm$0.000) & ($\pm$0.000)\\
            \multirow{2}{*}{MLP}
            & 0.787 & 0.614
            & \textbf{0.866} & 0.711
            & 0.856 & 0.688
            & 0.925 & 0.762
            & \textbf{0.334} & \textbf{0.247}\\
            & ($\pm$0.005) & ($\pm$0.005)
            & ($\pm$0.002) & ($\pm$0.004)
            & ($\pm$0.026) & ($\pm$0.015)
            & ($\pm$0.006) & ($\pm$0.008)
            & ($\pm$0.003) & ($\pm$0.001)\\
            \midrule
            \multirow{2}{*}{LR}
            & 0.630 & 0.441
            & 0.780 & 0.548
            & 0.830 & 0.594
            & 0.362 & 0.238
            & 0.247 & 0.169\\
            & ($\pm$0.001) & ($\pm$0.002)
            & ($\pm$0.003) & ($\pm$0.014)
            & ($\pm$0.031) & ($\pm$0.036)
            & ($\pm$0.012) & ($\pm$0.012)
            & ($\pm$0.001) & ($\pm$0.001)\\
            \multirow{2}{*}{CrossNet($t$=2)}
            & 0.717 & 0.525
            & 0.860 & 0.688
            & 0.860 & 0.655
            & 0.498 & 0.316
            & 0.310 & 0.211\\
            & ($\pm$0.001) & ($\pm$0.001)
            & ($\pm$0.005) & ($\pm$0.020)
            & ($\pm$0.057) & ($\pm$0.063)
            & ($\pm$0.008) & ($\pm$0.009)
            & ($\pm$0.001) & ($\pm$0.001)\\
            \multirow{2}{*}{EBM}
            & 0.802 & 0.612
            & 0.879 & 0.736
            & \textbf{0.905} & \textbf{0.726}
            & 0.906 & 0.737
            & 0.282 & 0.191\\
            & ($\pm$0.001) & ($\pm$0.002)
            & ($\pm$0.002) & ($\pm$0.005)
            & ($\pm$0.006) & ($\pm$0.007)
            & ($\pm$0.003) & ($\pm$0.004)
            & ($\pm$0.001) & ($\pm$0.001)\\
            \multirow{2}{*}{NAM}
            & 0.744 & 0.549
            & 0.781 & 0.544
            & 0.885 & 0.715
            & 0.694 & 0.476
            & 0.276 & 0.185\\
            & ($\pm$0.002) & ($\pm$0.003)
            & ($\pm$0.003) & ($\pm$0.009)
            & ($\pm$0.013) & ($\pm$0.009)
            & ($\pm$0.006) & ($\pm$0.005)
            & ($\pm$0.001) & ($\pm$0.002)\\
            \multirow{2}{*}{NodeGAM}
            & 0.734 & 0.540
            & 0.776 & 0.534
            & 0.897 & 0.732
            & 0.696 & 0.476
            & 0.275 & 0.188\\
            & ($\pm$0.004) & ($\pm$0.004)
            & ($\pm$0.003) & ($\pm$0.004)
            & ($\pm$0.008) & ($\pm$0.007)
            & ($\pm$0.005) & ($\pm$0.005)
            & ($\pm$0.001) & ($\pm$0.001)\\
            \multirow{2}{*}{NodeGA$^2$M}
            & 0.808 & 0.620
            & 0.879 & 0.732
            & 0.887 & 0.712
            & 0.911 & 0.740
            & 0.309 & 0.213\\
            & ($\pm$0.004) & ($\pm$0.005)
            & ($\pm$0.004) & ($\pm$0.010)
            & ($\pm$0.015) & ($\pm$0.024)
            & ($\pm$0.004) & ($\pm$0.004)
            & ($\pm$0.001) & ($\pm$0.001)\\
            \multirow{2}{*}{HONAM$^{*}$($t$=2)}
            & 0.807 & 0.621
            & 0.880 & \textbf{0.744}
            & 0.900 & 0.725
            & 0.920 & 0.752
            & \textbf{0.320} & 0.224\\
            & ($\pm$0.004) & ($\pm$0.004)
            & ($\pm$0.001) & ($\pm$0.003)
            & ($\pm$0.018) & ($\pm$0.017)
            & ($\pm$0.004) & ($\pm$0.008)
            & ($\pm$0.003) & ($\pm$0.003)\\
            \multirow{2}{*}{HONAM($t$=2)}
            & \textbf{0.810}$^*$ & \textbf{0.626}$^*$
            & \textbf{0.882}$^*$ & 0.742$^*$
            & 0.900 & 0.721
            & \textbf{0.925}$^*$ & \textbf{0.760}$^*$
            & \textbf{0.320}$^*$ & \textbf{0.226}$^*$\\
            & ($\pm$0.003) & ($\pm$0.003)
            & ($\pm$0.002) & ($\pm$0.012)
            & ($\pm$0.011) & ($\pm$0.009)
            & ($\pm$0.004) & ($\pm$0.006)
            & ($\pm$0.002) & ($\pm$0.004)\\
            \bottomrule
        \end{tabular}
    }
    
    \caption{Predictive performance comparison on regression tasks. $*$ indicates that the performance of HONAM is significantly better (p $<$ 0.05) than NAM.}
    
    \label{tab:regression}
    
\end{table*}
    \begin{table*}[!t]

    \centering

    \resizebox{.75\linewidth}{!}{
        \begin{tabular}{lcccccccccc}
            \toprule
            & \multicolumn{2}{c}{FICO}
            & \multicolumn{2}{c}{Credit}
            & \multicolumn{2}{c}{SUPPORT2}
            & \multicolumn{2}{c}{MIMIC-III}
            & \multicolumn{2}{c}{Click}\\
            & AUROC & AUPRC
            & AUROC & AUPRC
            & AUROC & AUPRC
            & AUROC & AUPRC
            & AUROC & AUPRC\\
            \midrule
            \multirow{2}{*}{XGBoost}
            & 0.767 & 0.744
            & \textbf{0.980} & \textbf{0.857}
            & 0.800 & 0.602
            & 0.791 & 0.342
            & \textbf{0.687} & \textbf{0.680}\\
            & ($\pm$0.003) & ($\pm$0.004)
            & ($\pm$0.006) & ($\pm$0.035)
            & ($\pm$0.014) & ($\pm$0.018)
            & ($\pm$0.009) & ($\pm$0.025)
            & ($\pm$0.001) & ($\pm$0.001)\\
            \multirow{2}{*}{MLP}
            & \textbf{0.770} & \textbf{0.749}
            & \textbf{0.980} & 0.803
            & \textbf{0.801} & \textbf{0.606}
            & \textbf{0.792} & \textbf{0.346}
            & 0.627 & 0.627\\
            & ($\pm$0.002) & ($\pm$0.003)
            & ($\pm$0.011) & ($\pm$0.051)
            & ($\pm$0.008) & ($\pm$0.013)
            & ($\pm$0.009) & ($\pm$0.025)
            & ($\pm$0.003) & ($\pm$0.002)\\
            \midrule
            \multirow{2}{*}{LR}
            & 0.753 & 0.723
            & 0.978 & 0.800
            & 0.783 & 0.577
            & 0.761 & 0.317
            & 0.617 & 0.620\\
            & ($\pm$0.001) & ($\pm$0.002)
            & ($\pm$0.012) & ($\pm$0.045)
            & ($\pm$0.009) & ($\pm$0.020)
            & ($\pm$0.012) & ($\pm$0.031)
            & ($\pm$0.002) & ($\pm$0.002)\\
            \multirow{2}{*}{CrossNet($t$=2)}
            & 0.770 & 0.752
            & 0.964 & 0.810
            & 0.803 & 0.612
            & 0.787 & 0.343
            & 0.619 & 0.625\\
            & ($\pm$0.002) & ($\pm$0.004)
            & ($\pm$0.024) & ($\pm$0.038)
            & ($\pm$0.009) & ($\pm$0.013)
            & ($\pm$0.005) & ($\pm$0.022)
            & ($\pm$0.001) & ($\pm$0.001)\\
            \multirow{2}{*}{EBM}
            & 0.700 & 0.625
            & 0.885 & 0.699
            & 0.674 & 0.434
            & 0.563 & 0.167
            & 0.592 & 0.551\\
            & ($\pm$0.002) & ($\pm$0.002)
            & ($\pm$0.023) & ($\pm$0.048)
            & ($\pm$0.017) & ($\pm$0.019)
            & ($\pm$0.012) & ($\pm$0.017)
            & ($\pm$0.001) & ($\pm$0.001)\\
            \multirow{2}{*}{NAM}
            & \textbf{0.783} & 0.760
            & 0.979 & 0.845
            & 0.815 & 0.626
            & 0.815 & 0.380
            & 0.655 & 0.652\\
            & ($\pm$0.003) & ($\pm$0.003)
            & ($\pm$0.014) & ($\pm$0.036)
            & ($\pm$0.014) & ($\pm$0.011)
            & ($\pm$0.005) & ($\pm$0.025)
            & ($\pm$0.002) & ($\pm$0.002)\\
            \multirow{2}{*}{NodeGAM}
            & 0.781 & \textbf{0.761}
            & 0.980 & \textbf{0.849}
            & 0.814 & 0.626
            & 0.813 & 0.375
            & 0.643 & 0.643\\
            & ($\pm$0.002) & ($\pm$0.002)
            & ($\pm$0.012) & ($\pm$0.040)
            & ($\pm$0.012) & ($\pm$0.010)
            & ($\pm$0.006) & ($\pm$0.017)
            & ($\pm$0.002) & ($\pm$0.001)\\
            \multirow{2}{*}{NodeGA$^2$M}
            & 0.780 & 0.760
            & \textbf{0.982} & 0.846
            & 0.812 & 0.624
            & 0.816 & 0.374
            & 0.641 & 0.639\\
            & ($\pm$0.002) & ($\pm$0.003)
            & ($\pm$0.011) & ($\pm$0.038)
            & ($\pm$0.012) & ($\pm$0.008)
            & ($\pm$0.009) & ($\pm$0.023)
            & ($\pm$0.002) & ($\pm$0.002)\\
            \multirow{2}{*}{HONAM$^{*}$($t$=2)}
            & \textbf{0.783} & \textbf{0.761}
            & \textbf{0.982} & 0.842
            & 0.819 & 0.633
            & 0.825 & 0.395
            & 0.667 & 0.663\\
            & ($\pm$0.003) & ($\pm$0.005)
            & ($\pm$0.009) & ($\pm$0.024)
            & ($\pm$0.009) & ($\pm$0.011)
            & ($\pm$0.003) & ($\pm$0.023)
            & ($\pm$0.004) & ($\pm$0.003)\\
            \multirow{2}{*}{HONAM($t$=2)}
            & 0.782 & 0.760
            & 0.981 & 0.838
            & \textbf{0.823} & \textbf{0.640}
            & \textbf{0.826}$^*$ & \textbf{0.399}
            & \textbf{0.670}$^*$ & \textbf{0.664}$^*$\\
            & ($\pm$0.002) & ($\pm$0.004)
            & ($\pm$0.012) & ($\pm$0.026)
            & ($\pm$0.011) & ($\pm$0.009)
            & ($\pm$0.005) & ($\pm$0.024)
            & ($\pm$0.002) & ($\pm$0.003)\\
            \bottomrule
        \end{tabular}
    }
    
    \caption{Predictive performance comparison on classification tasks. $*$ indicates that the performance of HONAM is significantly better (p $<$ 0.05) than NAM.}
    
    \label{tab:classification}
    
\end{table*}

    We conducted our experiments using the following 10 publicly available datasets:
        The California Housing dataset \cite{california} contains information from the 1990 California census and is used to predict median house values in California districts.
        The Insurance dataset \cite{insurance} contains personal health information and is used to predict individual medical costs billed by health insurance.
        The House Prices dataset \cite{house} contains information on various housing attributes, such as location and number of rooms, and is used to predict the selling price of houses.
        The Bikeshare dataset \cite{bike} contains counts of rented bikes along with associated weather and seasonal features.
        The Year dataset \cite{year} contains features of songs spanning from 1922 to 2011 and aims to predict each song’s release year.
        The FICO dataset \cite{fico} comprises data from consumers requesting credit lines and aims to predict whether individuals with opened credit accounts experienced payment delays exceeding 90 days within the past 24 months.
        The Credit dataset \cite{Dal2015} includes de-identified features collected for credit fraud detection. For further details, please refer to the original source \cite{Dal2015}.
        The Study to Understand Prognoses, Preferences, Outcomes, and Risks of Treatment 2 (SUPPORT2) dataset \cite{support2} contains patient information collected to investigate prognosis preference outcomes and treatment risks.
        The MIMIC-III dataset \cite{johnson2016mimic} are large-scale databases containing hospitalization records, prescription information, etc. Although these datasets support multiple tasks, our primary focus is on patient mortality prediction.
        The Click dataset \cite{click} contains advertising data used to predict a user’s likelihood of clicking an advertisement. Following previous work \cite{popov2019neural}, we extracted 500,000 positive and negative samples for our experiment.
    Table~\ref{tab:dataset} presents the statistical information of the datasets.

    \subsection{Experimental Setup}

    For the Year and Click datasets, we utilized the predefined training, validation, and test sets from previous work \cite{popov2019neural}. For the remaining datasets, we randomly divided the data into training, validation, and test sets with proportions of 60\%, 20\%, and 20\%, respectively, using five distinct random seeds.

    We applied ordinal encoding to categorical features to reduce memory usage. Continuous features were standardized to have zero mean and unit variance. Subsequently, we applied quantile transformation to all features, adding a small amount of Gaussian noise during this process. This method ensures that the mean and standard deviation remain close to 0 and 1, respectively \cite{chang2022node}.

    The experiments were conducted on a machine equipped with an Intel i7-8700 CPU, NVIDIA GeForce RTX 3090 Ti GPU, and 64 GB of RAM.

    \begin{table*}[!t]

    \centering
    
    \resizebox{.75\linewidth}{!}{
        \begin{tabular}{lcccccccccc}
            \toprule
            & \multicolumn{2}{c}{California Housing}
            & \multicolumn{2}{c}{Insurance}
            & \multicolumn{2}{c}{House Prices}
            & \multicolumn{2}{c}{Bikeshare}
            & \multicolumn{2}{c}{Year}\\
            & R-squared & R-absolute
            & R-squared & R-absolute
            & R-squared & R-absolute
            & R-squared & R-absolute
            & R-squared & R-absolute\\
            \midrule
            \multirow{2}{*}{CrossNet($t$=2)}
            & 0.717
            & 0.525
            & \textbf{0.860}
            & 0.688
            & 0.885
            & 0.683
            & 0.498
            & 0.316
            & 0.310
            & 0.211 \\
            & ($\pm$0.001)
            & ($\pm$0.001)
            & ($\pm$0.005)
            & ($\pm$0.020)
            & ($\pm$0.011)
            & ($\pm$0.003)
            & ($\pm$0.008)
            & ($\pm$0.009)
            & ($\pm$0.001)
            & ($\pm$0.001) \\
            \multirow{2}{*}{CrossNet($t$=3)}
            & 0.739
            & 0.549
            & 0.854
            & \textbf{0.691}
            & \textbf{0.903}
            & \textbf{0.720}
            & 0.590
            & 0.403
            & 0.324
            & 0.228 \\
            & ($\pm$0.004)
            & ($\pm$0.004)
            & ($\pm$0.007)
            & ($\pm$0.020)
            & ($\pm$0.007)
            & ($\pm$0.007)
            & ($\pm$0.017)
            & ($\pm$0.009)
            & ($\pm$0.001)
            & ($\pm$0.002) \\
            \multirow{2}{*}{CrossNet($t$=4)}
            & \textbf{0.741}
            & \textbf{0.562}
            & 0.855
            & 0.690
            & \textbf{0.903}
            & 0.718
            & \textbf{0.645}
            & \textbf{0.453}
            & \textbf{0.327}
            & \textbf{0.236} \\
            & ($\pm$0.007)
            & ($\pm$0.006)
            & ($\pm$0.010)
            & ($\pm$0.007)
            & ($\pm$0.007)
            & ($\pm$0.008)
            & ($\pm$0.013)
            & ($\pm$0.011)
            & ($\pm$0.002)
            & ($\pm$0.002) \\
            \midrule
            \multirow{2}{*}{HONAM$^{*}$($t$=2)}
            & 0.807
            & 0.621
            & 0.880
            & \textbf{0.744}
            & 0.900
            & 0.725
            & 0.920
            & 0.752
            & 0.320
            & 0.224 \\
            & ($\pm$0.004)
            & ($\pm$0.004)
            & ($\pm$0.001)
            & ($\pm$0.003)
            & ($\pm$0.018)
            & ($\pm$0.017)
            & ($\pm$0.004)
            & ($\pm$0.008)
            & ($\pm$0.003)
            & ($\pm$0.003) \\
            \multirow{2}{*}{HONAM$^{*}$($t$=3)}
            & \textbf{0.810}
            & \textbf{0.629}
            & \textbf{0.882}
            & 0.743
            & \textbf{0.905}
            & \textbf{0.730}
            & 0.945
            & 0.806
            & 0.329
            & 0.235 \\
            & ($\pm$0.008)
            & ($\pm$0.009)
            & ($\pm$0.001)
            & ($\pm$0.005)
            & ($\pm$0.020)
            & ($\pm$0.022)
            & ($\pm$0.005)
            & ($\pm$0.010)
            & ($\pm$0.004)
            & ($\pm$0.006) \\
            \multirow{2}{*}{HONAM$^{*}$($t$=4)}
            & 0.804
            & 0.624
            & 0.881
            & 0.741
            & 0.904
            & 0.728
            & \textbf{0.949}
            & \textbf{0.816}
            & \textbf{0.331}
            & \textbf{0.243} \\
            & ($\pm$0.005)
            & ($\pm$0.004)
            & ($\pm$0.002)
            & ($\pm$0.006)
            & ($\pm$0.025)
            & ($\pm$0.034)
            & ($\pm$0.003)
            & ($\pm$0.004)
            & ($\pm$0.005)
            & ($\pm$0.008) \\
            \midrule
            \multirow{2}{*}{HONAM($t$=2)}
            & \textbf{0.810}
            & \textbf{0.626}
            & \textbf{0.882}
            & \textbf{0.742}
            & 0.900
            & 0.721
            & 0.925
            & 0.760
            & 0.320
            & 0.226 \\
            & ($\pm$0.003)
            & ($\pm$0.003)
            & ($\pm$0.002)
            & ($\pm$0.012)
            & ($\pm$0.011)
            & ($\pm$0.009)
            & ($\pm$0.004)
            & ($\pm$0.006)
            & ($\pm$0.002)
            & ($\pm$0.004) \\
            \multirow{2}{*}{HONAM($t$=3)}
            & 0.804
            & 0.623
            & 0.881
            & 0.741
            & 0.904
            & 0.728
            & \textbf{0.952}
            & 0.817
            & \textbf{0.327}
            & \textbf{0.237} \\
            & ($\pm$0.004)
            & ($\pm$0.001)
            & ($\pm$0.002)
            & ($\pm$0.004)
            & ($\pm$0.016)
            & ($\pm$0.004)
            & ($\pm$0.003)
            & ($\pm$0.004)
            & ($\pm$0.005)
            & ($\pm$0.006) \\
            \multirow{2}{*}{HONAM($t$=4)}
            & 0.805
            & 0.619
            & 0.881
            & 0.740
            & \textbf{0.909}
            & \textbf{0.735}
            & \textbf{0.952}
            & \textbf{0.819}
            & 0.326
            & \textbf{0.237} \\

            & ($\pm$0.003)
            & ($\pm$0.004)
            & ($\pm$0.002)
            & ($\pm$0.005)
            & ($\pm$0.015)
            & ($\pm$0.011)
            & ($\pm$0.003)
            & ($\pm$0.004)
            & ($\pm$0.002)
            & ($\pm$0.005) \\
            \bottomrule
        \end{tabular}
    }
    
    \caption{Interaction ablation study on regression tasks}
    
    \label{tab:regression_ablation}
    
\end{table*}
    \begin{table*}[!t]

    \centering

    \resizebox{.75\linewidth}{!}{
        \begin{tabular}{lcccccccccc}
            \toprule
            & \multicolumn{2}{c}{FICO}
            & \multicolumn{2}{c}{Credit}
            & \multicolumn{2}{c}{SUPPORT2}
            & \multicolumn{2}{c}{MIMIC-III}
            & \multicolumn{2}{c}{Click}\\
            & AUROC & AUPRC
            & AUROC & AUPRC
            & AUROC & AUPRC
            & AUROC & AUPRC
            & AUROC & AUPRC\\
            \midrule
            \multirow{2}{*}{CrossNet($t$=2)}
            & 0.770
            & 0.752
            & 0.964
            & 0.810
            & \textbf{0.803}
            & \textbf{0.612}
            & \textbf{0.787}
            & \textbf{0.343}
            & \textbf{0.619}
            & \textbf{0.625} \\
            & ($\pm$0.002)
            & ($\pm$0.004)
            & ($\pm$0.024)
            & ($\pm$0.038)
            & ($\pm$0.009)
            & ($\pm$0.013)
            & ($\pm$0.005)
            & ($\pm$0.022)
            & ($\pm$0.001)
            & ($\pm$0.001) \\
            \multirow{2}{*}{CrossNet($t$=3)}
            & \textbf{0.772}
            & 0.754
            & \textbf{0.969}
            & 0.824
            & 0.794
            & 0.601
            & 0.768
            & 0.341
            & 0.616
            & 0.616 \\
            & ($\pm$0.001)
            & ($\pm$0.002)
            & ($\pm$0.016)
            & ($\pm$0.042)
            & ($\pm$0.013)
            & ($\pm$0.019)
            & ($\pm$0.010)
            & ($\pm$0.022)
            & ($\pm$0.003)
            & ($\pm$0.003) \\
            \multirow{2}{*}{CrossNet($t$=4)}
            & \textbf{0.772}
            & \textbf{0.755}
            & 0.965
            & \textbf{0.838}
            & 0.798
            & 0.605
            & 0.775
            & 0.342
            & 0.617
            & 0.619 \\
            & ($\pm$0.002)
            & ($\pm$0.003)
            & ($\pm$0.024)
            & ($\pm$0.033)
            & ($\pm$0.007)
            & ($\pm$0.010)
            & ($\pm$0.009)
            & ($\pm$0.027)
            & ($\pm$0.003)
            & ($\pm$0.003) \\
            \midrule
            \multirow{2}{*}{HONAM$^{*}$($t$=2)}
            & \textbf{0.783}
            & \textbf{0.761}
            & 0.982
            & 0.842
            & \textbf{0.819}
            & \textbf{0.633}
            & \textbf{0.825}
            & \textbf{0.395}
            & \textbf{0.667}
            & \textbf{0.663} \\
            & ($\pm$0.003)
            & ($\pm$0.005)
            & ($\pm$0.009)
            & ($\pm$0.024)
            & ($\pm$0.009)
            & ($\pm$0.011)
            & ($\pm$0.003)
            & ($\pm$0.023)
            & ($\pm$0.004)
            & ($\pm$0.003) \\
            \multirow{2}{*}{HONAM$^{*}$($t$=3)}
            & 0.781
            & 0.759
            & \textbf{0.987}
            & \textbf{0.853}
            & \textbf{0.819}
            & 0.632
            & 0.824
            & 0.388
            & \textbf{0.667}
            & 0.656 \\
            & ($\pm$0.002)
            & ($\pm$0.003)
            & ($\pm$0.006)
            & ($\pm$0.031)
            & ($\pm$0.008)
            & ($\pm$0.014)
            & ($\pm$0.002)
            & ($\pm$0.023)
            & ($\pm$0.003)
            & ($\pm$0.003)\\
            \multirow{2}{*}{HONAM$^{*}$($t$=4)}
            & 0.780
            & 0.758
            & 0.982
            & 0.851
            & \textbf{0.819}
            & \textbf{0.633}
            & 0.822
            & 0.386
            & 0.664
            & 0.657 \\
            & ($\pm$0.001)
            & ($\pm$0.002)
            & ($\pm$0.017)
            & ($\pm$0.061)
            & ($\pm$0.010)
            & ($\pm$0.012)
            & ($\pm$0.005)
            & ($\pm$0.022)
            & ($\pm$0.006)
            & ($\pm$0.005)\\
            \midrule
            \multirow{2}{*}{HONAM($t$=2)}
            & \textbf{0.782}
            & \textbf{0.760}
            & 0.981
            & 0.838
            & \textbf{0.823}
            & \textbf{0.640}
            & \textbf{0.826}
            & \textbf{0.399}
            & \textbf{0.670}
            & \textbf{0.664} \\
            & ($\pm$0.002)
            & ($\pm$0.004)
            & ($\pm$0.012)
            & ($\pm$0.026)
            & ($\pm$0.011)
            & ($\pm$0.009)
            & ($\pm$0.005)
            & ($\pm$0.024)
            & ($\pm$0.002)
            & ($\pm$0.003) \\
            \multirow{2}{*}{HONAM($t$=3)}
            & \textbf{0.782}
            & \textbf{0.760}
            & 0.981
            & \textbf{0.843}
            & 0.820
            & 0.637
            & \textbf{0.826}
            & 0.397
            & 0.669
            & 0.663 \\
            & ($\pm$0.002)
            & ($\pm$0.004)
            & ($\pm$0.009)
            & ($\pm$0.031)
            & ($\pm$0.010)
            & ($\pm$0.014)
            & ($\pm$0.004)
            & ($\pm$0.022)
            & ($\pm$0.002)
            & ($\pm$0.002) \\
            \multirow{2}{*}{HONAM($t$=4)}
            & 0.780
            & 0.758
            & \textbf{0.983}
            & 0.840
            & 0.819
            & 0.636
            & 0.824
            & 0.390
            & 0.668
            & 0.662 \\
            & ($\pm$0.002)
            & ($\pm$0.004)
            & ($\pm$0.009)
            & ($\pm$0.035)
            & ($\pm$0.010)
            & ($\pm$0.013)
            & ($\pm$0.004)
            & ($\pm$0.019)
            & ($\pm$0.005)
            & ($\pm$0.005) \\
            \bottomrule
        \end{tabular}
    }
    
    \caption{Interaction ablation study on classification tasks}
    
    \label{tab:classification_ablation}
    
\end{table*}

    \subsection{Hyperparameters}

    In this study, we evaluated various models, including linear/logistic regression, CrossNet \cite{wang2017deep}, XGBoost \cite{Chen2016}, MLP, EBM \cite{nori2019interpret}, NAM \cite{agarwal2020neural}, NodeGAM \cite{chang2021interpretable}, NodeGA$^{\text{2}}$M \cite{chang2021interpretable}, and our proposed HONAM. We used open-source implementations for XGBoost and EBM, while the other methods were implemented in PyTorch. For MLP, NAM, and HONAM, the network architecture consisted of three hidden layers with [32, 64, 32] hidden units and LeakyReLU activation. The number of units in CrossNet was set to 32. All PyTorch models were trained for 1,000 epochs with a learning rate of 0.001, using a batch size set to approximately 1\% of the dataset size to optimize training time. The model achieving the best validation performance was selected for final evaluation. For NodeGAM and NodeGA$^{\text{2}}$M, we used the recommended hyperparameter setting suggested in the original study \cite{chang2021interpretable}. For XGBoost, we used 1,000 boosting rounds to ensure convergence, with a learning rate (eta) of 0.3. For EBM, we set the number of boosting rounds to 20,000, with both inner and outer bagging parameters set to 8, and used a learning rate of 0.01. All experiments were conducted using five random seeds, and we report the mean scores along with their standard deviations.

    \subsection{Evaluation Metrics}

    The R-squared score is a widely used metric for regression tasks but relies solely on the mean squared error, offering a limited perspective on regression performance. To address this limitation, we propose the R-absolute score, a novel scaled regression metric based on the mean absolute error, providing a complementary evaluation measure. The R-absolute score is defined as follows:
    \begin{equation}
    R\text{-}absolute = 1 - \frac{ \sum_{i=1}^{N} \left\vert y_{i} - \hat{y}_{i} \right\vert }{ \sum_{i=1}^{N} \left\vert y_{i} - \bar{y} \right\vert },
\end{equation}\noindent
    where $N$ denotes the number of data samples. For regression tasks, we employed both R-squared and R-absolute scores as evaluation metrics. For classification tasks, we utilized the area under the receiver operating characteristic curve (AUROC) and the area under the precision-recall curve (AUPRC).

    \begin{figure*}[!t]

    \centering
    
    \begin{subfigure}{0.45\linewidth}
        \centering
        \includegraphics[height=7cm]{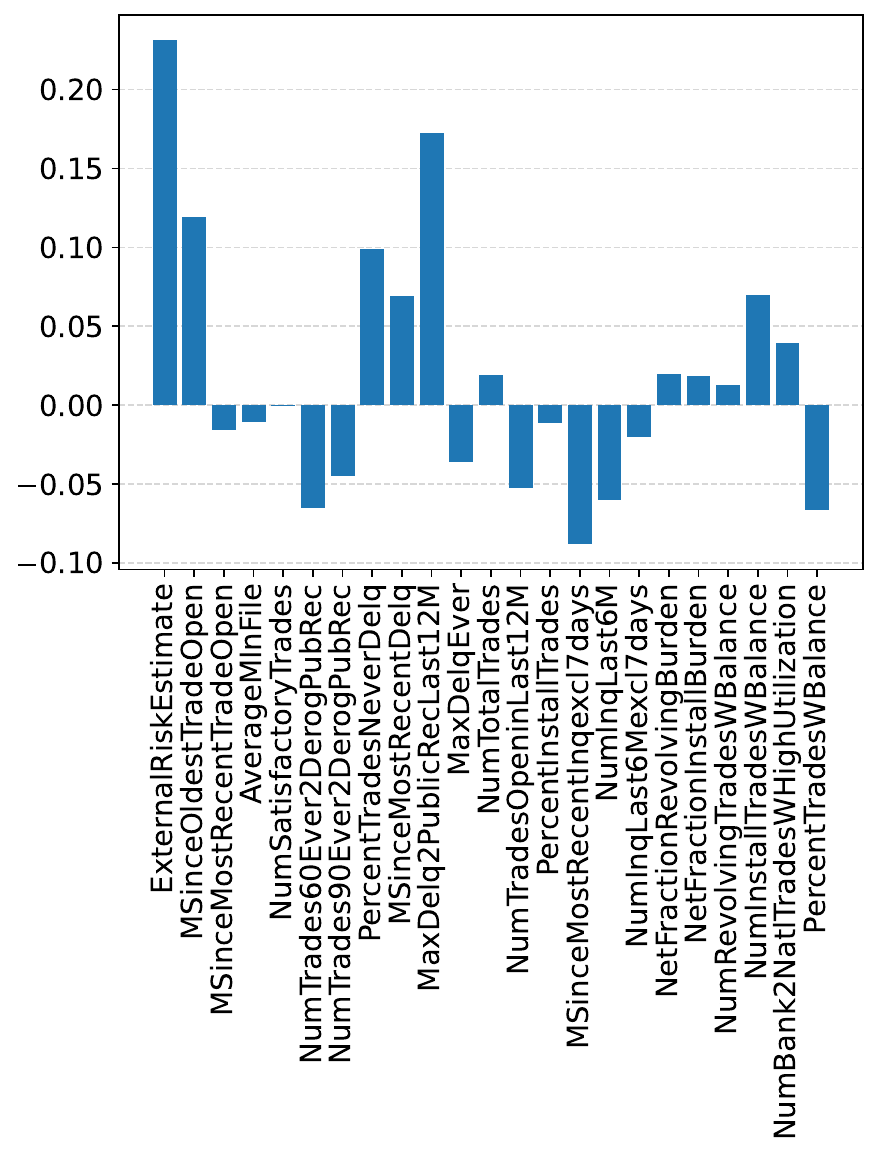}
        \caption{Effects of first-order features. Features represented by longer bars have a more substantial impact on the prediction than features represented by shorter bars.}
        \label{fig:local_first}
    \end{subfigure}
    \begin{subfigure}{0.45\linewidth}
        \centering
        \includegraphics[height=7cm]{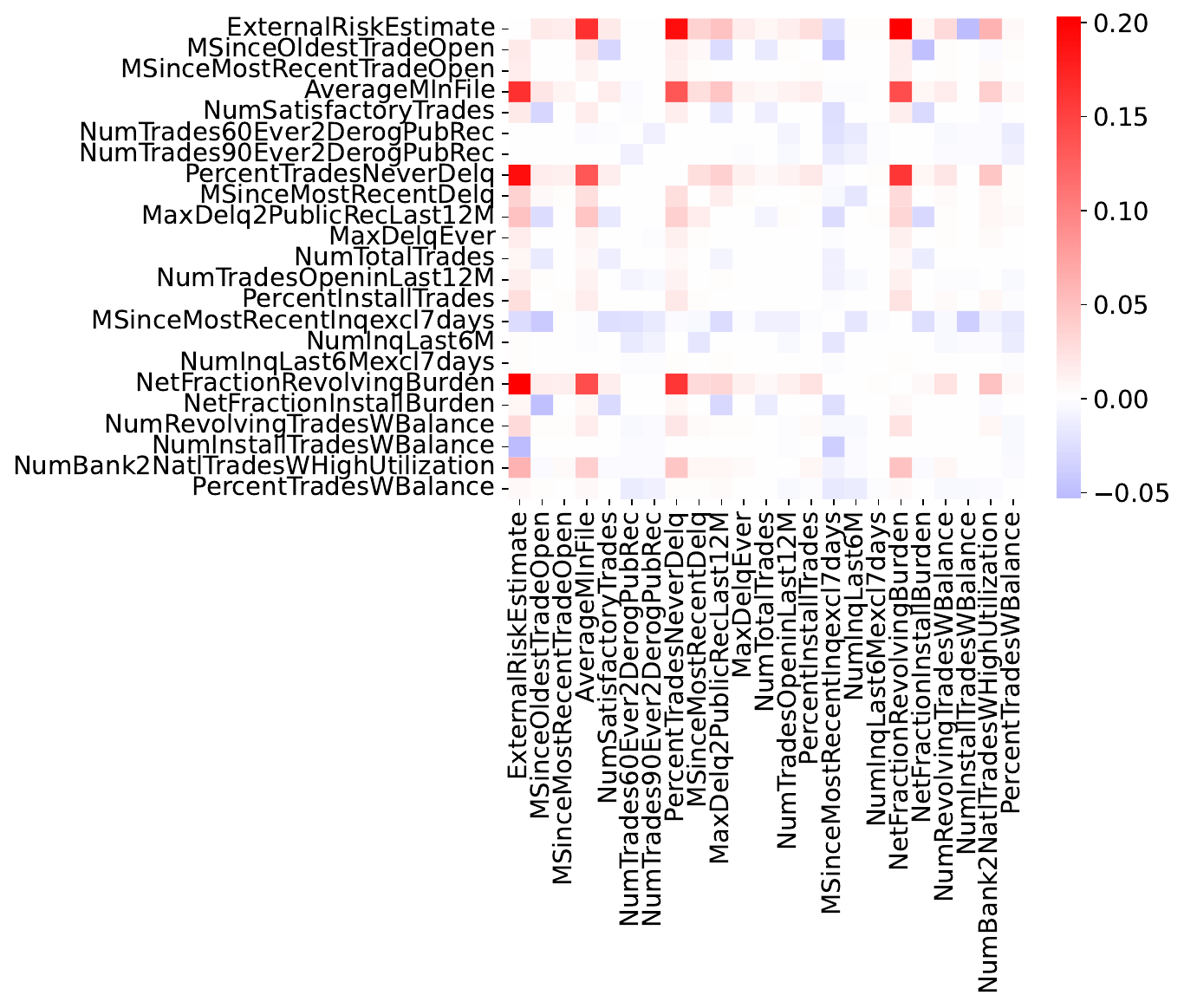}
        \caption{Effects of second-order features. Red and blue cells indicate features that have positive and negative effects, respectively.}
        \label{fig:local_second}
    \end{subfigure}
    
    \caption{Local interpretations for the FICO dataset.}
    
\end{figure*}
    \begin{figure*}[!t]

    \centering
    
    \begin{subfigure}{0.45\linewidth}
        \centering
        \includegraphics[width=6cm]{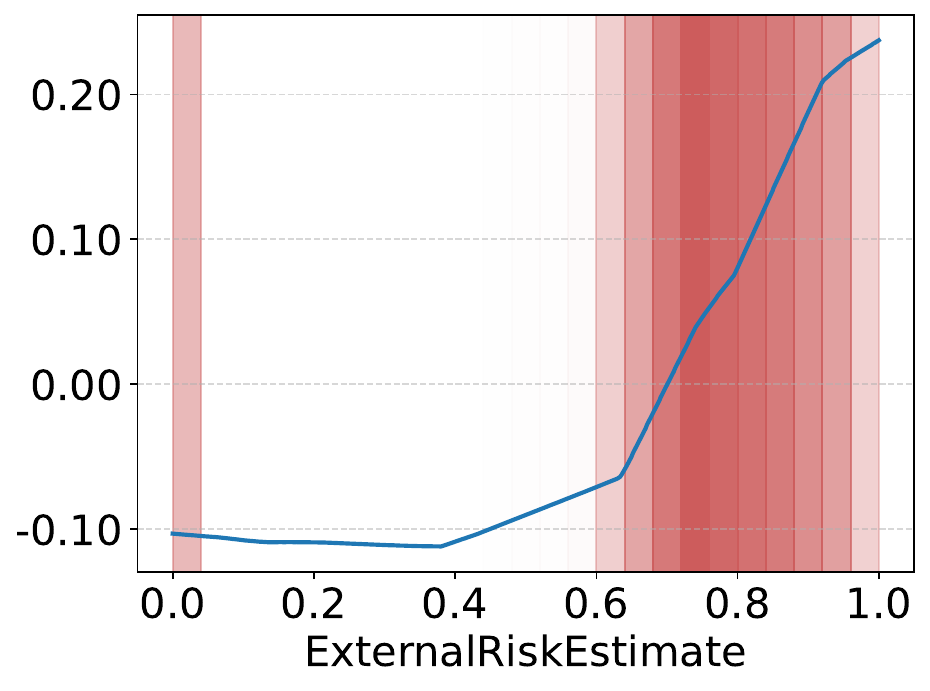}
        \caption{Effects of a first-order feature, \textit{ExternalRiskEstimate}. The red bars indicate the densities of training data.}
        \label{fig:global_first}
    \end{subfigure}
    \begin{subfigure}{0.45\linewidth}
        \centering
        \includegraphics[width=6cm]{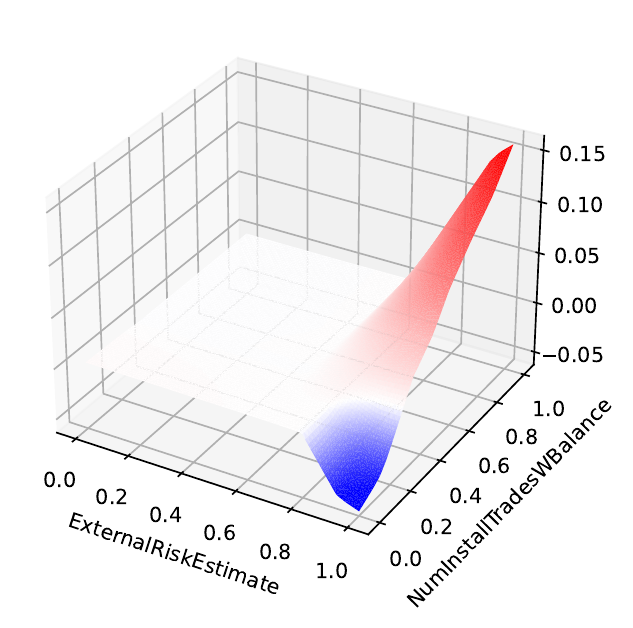}
        \caption{Effects of a second-order feature, \textit{ExternalRiskEstimate $\times$ NumInstallTradesWBalance}.}
        \label{fig:global_second}
    \end{subfigure}
    
    \caption{Global interpretations for the FICO dataset}	
    
\end{figure*}

    \subsection{Effectiveness of Feature Interactions}

    Although the primary focus of this study is the interpretability of HONAM, we also demonstrate the effectiveness of feature interaction modeling in enhancing predictive performance. To this end, we compared HONAM with several machine learning models across various regression and classification datasets. Table~\ref{tab:regression} and Table~\ref{tab:classification} show the performances of the experimental models for the regression and classification tasks. In these tables, \textit{HONAM$^{*}$} denotes HONAM combined with CrossNet, while \textit{HONAM} refers to HONAM combined with the proposed interaction module. The experimental results indicate that HONAM outperforms EBM, NAM, and NodeGAM across both regression and classification tasks, underscoring the effectiveness of feature interaction learning. Furthermore, HONAM achieves superior results compared to NodeGA$^{\text{2}}$M, highlighting the advantages of its fully neural network-based approach over the tree-based method used in NodeGAM. HONAM also demonstrates comparable or even superior performance compared to black-box models, such as MLP and XGBoost. Interestingly, certain GAM-based models occasionally outperform MLP on specific datasets, a phenomenon can be attributed to differences in feature distributions. Handling input features with distinct distributions using shared trainable parameters may disturb training \cite{Yan2022}. In contrast, GAM-family models alleviate this issue by employing separate trainable parameters for each feature.

    We conducted an ablation study on feature interactions using 2nd-, 3rd-, and 4th-order CrossNets and HONAMs to evaluate the impact of increasing the interaction order on predictive performance. Table~\ref{tab:regression_ablation} and Table~\ref{tab:classification_ablation} present the results for regression and classification tasks, respectively. For the House Prices, Bikeshare, and Year datasets, predictive performance improved with higher-order feature interactions. However, significant improvements were not observed for the remaining datasets, possibly due to the absence of meaningful higher-order interactions in those datasets or limitations in model capacity. High-order CrossNet and HONAM require larger model sizes than their low-order counterparts due to the larger number of unique interactions in high-order features. Nevertheless, even without performance gains, high-order interactions remain beneficial for interpretability, offering deeper insights into complex feature relationships.

    \subsection{Interpretations}

    In HONAM, each input feature is processed independently by its corresponding MLP, allowing the output of each MLP to directly represent the individual contribution of that feature to the prediction. Local interpretations, obtained through forward propagation, reveal how individual features contribute to a specific output. Aggregating these local interpretations enables global insights into HONAM’s overall behavior. In this study, we visualized local interpretations for 1st- and 2nd-order feature interactions and provided global interpretations for a comprehensive understanding of HONAM’s behavior. Note that HONAM also supports interpretations of higher-order features (e.g., 3rd- and 4th-order interactions), a capability not available in other GAM-based models such as NAM, NodeGAM, and NodeGA$^2$M.

    Fig.~\ref{fig:local_first} and Fig.~\ref{fig:local_second} depict visualizations of local interpretations for 1st- and 2nd-order features in the FICO dataset, respectively. These visualizations show the actual contribution of individual features toward the model’s predictions. Since the FICO dataset is a binary classification task, features with positive contributions increase the probability, while those with negative contributions decrease the probability. In addition, Fig.~\ref{fig:local_second} provides a heat map visualization of 2nd-order feature interactions, where red cells indicate positive influences and blue cells represent negative influences.

    Fig.~\ref{fig:global_first} and Fig.~\ref{fig:global_second} illustrate the global interpretations of the 1st-order feature \textit{ExternalRiskEstimate} and the 2nd-order feature \textit{ExternalRiskEstimate $\times$ NumInstallTradesWBalance} in the FICO dataset, respectively. In Fig.~\ref{fig:global_first}, the red bars indicate the densities of training samples. In the 1st-order interpretation, we observe a tendency for an increase in \textit{ExternalRiskEstimate} to contribute positively. Interestingly, the 2nd-order interpretation exhibits a different pattern. Despite a high value for \textit{ExternalRiskEstimate}, it have a negative impact if \textit{NumInstallTradesWBalance} is low. This tendency cannot be captured in 1st-order interactions and can only be observed in 2nd-order interactions. This observation demonstrates HONAM’s ability to offer richer and more detailed interpretations than NAM, emphasizing the necessity of modeling higher-order feature interactions in interpretable AI.

\section{Limitations \& Future Works}

The primary limitation of NAM-family models, including HONAM, is their slow inference time. These models require separate neural networks for each input feature, which enhances interpretability but linearly increases computational time and model size to the number of features. Fortunately, GPU parallelization using block-sparse layer or grouped convolution layer can alleviate the computational time. Moreover, various methods have been proposed to enhance the scalability of GAMs. For example, \cite{chang2022node} introduced a neural oblivious tree-based GAM, which leverages oblivious trees to reduce the number of feature functions and computational time compared to neural network-based approaches. \cite{filip_radenovic_neural_2022} presented a basis network-based GAM, where input features share a fixed number of basis functions rather than using separate functions for each feature. Additionally, \cite{xu_sparse_2023} proposed a sparse NAM variant employing group LASSO to identify and eliminate insignificant features, further improving model efficiency.

The proposed recursive interaction module generates all possible feature combinations, but this increases HONAM’s execution time and introduces the possibility of noisy interactions \cite{cheng2020adaptive}. Adaptive feature interaction strategies could address this issue. For instance, as illustrated in Fig.~\ref{fig:local_second}, most cells appear white, indicating negligible contributions to the predictions. Such insignificant interactions could be discarded, leading to a more efficient and robust model. Future work should investigate methods for adaptively identifying only meaningful feature interactions in interpretable models.

\section{Conclusion}

In this study, we introduced HONAM, a novel interpretable machine learning model capable of capturing feature interactions of arbitrary orders. Through comprehensive experiments, we demonstrated that our proposed interaction method significantly improves predictive performance. By visualizing both local and global interpretations for 1st- and 2nd-order feature interactions, we highlighted the importance of modeling higher-order interactions to enhance interpretability. Given its effectiveness and transparency, we anticipate HONAM will gain widespread adoption across diverse domains.

\appendix

\subsection{Proof of Proposition 1}\label{sec:proof1}

\textbf{Definition A.1}. The sum of the $t^\text{th}$-order feature interactions, excluding powered terms of features, is defined as follows:
\begin{equation}
    \begin{split}
        g_{t} \left( \mathbf{Z} \right) &= \sum_{j_{1} = 1}^{m} \sum_{j_{2} = j_{1} + 1}^{m} \ldots \sum_{j_{t} = j_{t-1} + 1}^{m} \bigodot_{l=1}^{t} \mathbf{z}_{j_{l}}\\
        &=: \sum_{j_{1} > \ldots > j_{t}}^{m} \bigodot_{l=1}^{t} \mathbf{z}_{j_{l}}.
    \end{split}
\end{equation}

To prove \textbf{Proposition 1}, we show that the proposed recursive method satisfies \textbf{Definition A.1} using mathematical induction. First, we show that \textbf{Proposition 1} holds for the interaction order $t = 2$:
\begin{equation}
    \begin{split}
        g_{2} \left( \mathbf{Z} \right)
        &= \frac{1}{2} \sum_{i=1}^{2} \left( \left( -1 \right)^{i+1} g_{1} \left( \mathbf{Z}^{i} \right)  g_{t-i} \left( \mathbf{Z} \right) \right) \\
        &= \frac{1}{2} \left( g_{1} \left( \mathbf{Z} \right) \odot g_{1} \left( \mathbf{Z} \right) - g_{1} \left( \mathbf{Z}^{2} \right) \odot g_{0} \left( \mathbf{Z} \right) \right) \\
        &= \frac{1}{2} \left( \sum_{j=1}^{m} \mathbf{z}_{j} \odot \sum_{j=1}^{m} \mathbf{z}_{j} - \sum_{j=1}^{m} \mathbf{z}^{2}_{j} \right) \\
        &= \sum_{j_{1}=1}^{m} \sum_{j_{2}={j_{1}+1}}^{m} \left( \mathbf{z}_{j_{1}} \odot \mathbf{z}_{j_{2}} \right),
    \end{split}
\end{equation}\noindent
which satisfies \textbf{Definition A.1}. Then, we assume that \textbf{Proposition 1} holds for interaction order $t = k - 1$, that is,
\begin{equation}
    \begin{split}
        g_{k-1} \left( \mathbf{Z} \right)
        &= \frac{1}{k-1} \sum_{i=1}^{k-1} \left( \left( -1 \right)^{i+1} g_{1} \left( \mathbf{Z}^i \right) \odot g_{k-i-1} \left( \mathbf{Z} \right) \right) \\
        &= \sum_{j_{1} > \ldots > j_{k-1}}^{m} \bigodot_{l=1}^{k-1} \mathbf{z}_{j_{l}}.
    \end{split}
    \label{eq:induction}
\end{equation}\noindent
Next, we show that \textbf{Proposition 1} holds for interaction order $t = k$. Multiplying the sum of $(k-1)^\text{th}$-order interactions by the sum of $1^\text{st}$-order interactions is defined as follows:
\begin{equation}
    \begin{split}
        & \sum_{j=1}^{m} \mathbf{z}_{j} \odot g_{k-1} \left( \mathbf{Z} \right)=\sum_{j=1}^{m} \mathbf{z}_{j} \odot \sum_{j_{1} > \ldots > j_{k-1}}^{m} \bigodot_{l=1}^{k-1} \mathbf{z}_{j_{l}} \\
        &= k \sum_{j_{1} > \ldots >j_{k}}^{m} \bigodot_{l=1}^{k} \mathbf{z}_{j_{l}} +
        \sum_{j=1}^{m} \left( \mathbf{z}_{j}^{2} \odot \underset{j \neq j_{*}}{ \sum_{j_{1} > \ldots > j_{k-2}}^{m} } \bigodot_{l=1}^{k-2} \mathbf{z}_{j_{l}} \right) \\
        &= k \sum_{j_{1} > \ldots > j_{k}}^{m} \bigodot_{l=1}^{k} \mathbf{z}_{j_{l}} + \sum_{j=1}^{m} \mathbf{z}_{j}^{2} \odot \sum_{j_{1} > \ldots > j_{k-2}}^{m} \bigodot_{l=1}^{k-2} \mathbf{z}_{j_{l}} \\
        & \hspace{73pt} - \sum_{j=1}^{m} \left( \mathbf{z}_{j}^{3} \odot \underset{j \neq j_{*}}{ \sum_{j_{1} > \ldots > j_{k-3}}^{m} } \bigodot_{l=1}^{k-3} \mathbf{z}_{j_{l}} \right) \\
        &= k \sum_{j_{1} > \ldots > j_{k}}^{m} \bigodot_{l=1}^{k} \mathbf{z}_{j_{l}} + \sum_{j=1}^{m} \mathbf{z}_{j}^{2} \odot \sum_{j_{1} > \ldots > j_{k-2}}^{m} \bigodot_{l=1}^{k-2} \mathbf{z}_{j_{l}} \\
        & \hspace{58pt} - \sum_{j=1}^{m} \mathbf{z}_{j}^{3} \odot \sum_{j_{1} > \ldots > j_{k-3}}^{m} \bigodot_{l=1}^{k-3} \mathbf{z}_{j_{l}} + \ldots \\
        &= k \sum_{j_{1} > \ldots > j_{k}}^{m} \bigodot_{l=1}^{k} \mathbf{z}_{j_{l}}\\
        & \hspace{18pt} + \sum_{i=2}^{k} \left( \left( -\mathbf{1} \right)^{i} \odot \sum_{j=1}^{m} \mathbf{z}_{j}^{i} \odot \sum_{j_{1} > \ldots > j_{k-i}}^{m} \bigodot_{l=1}^{k-i} \mathbf{z}_{j_{l}} \right).
    \end{split}
    \label{eq:proof3}
\end{equation}\noindent
Therefore, eliminating powered terms from \eqref{eq:proof3} is defined as follows:
\begin{equation}
    \begin{split}
        & \sum_{j=1}^{m} \mathbf{z}_{j} \odot \sum_{i_{1} > \ldots > i_{k-1}}^{m} \bigodot_{l=1}^{k-1} \mathbf{z}_{i_{l}}\\
        & \hspace{20pt} - \sum_{i=2}^{k} \left( \left( -\mathbf{1} \right)^{i} \odot \sum_{j=1}^{m} \mathbf{z}_{j}^{i} \odot \sum_{j_{1} > \ldots > j_{k-i}}^{m} \bigodot_{l=1}^{k-i} \mathbf{z}_{j_{l}} \right) \\
        &= \sum_{i=1}^{k} \left( \left( -\mathbf{1} \right)^{i+1} \odot \sum_{j=1}^{m} \mathbf{z}_{j}^{i} \odot \sum_{j_{1} > \ldots > j_{k-i}}^{m} \bigodot_{l=1}^{k-i} \mathbf{z}_{j_{l}} \right) \\
        &= k \sum_{j_{1} > \ldots > j_{k}}^{m} \bigodot_{l=1}^{k} \mathbf{z}_{j_{l}}.
    \end{split}
\end{equation}\noindent
Subsequently, by eliminating duplicate interactions through division by $k$ and applying \eqref{eq:induction}, we obtain the following recursive form for interactions:
\begin{equation}
    \begin{split}
        g_{k} \left( \mathbf{Z} \right) &= \sum_{j_{1} > \ldots > j_{k}}^{m} \bigodot_{l=1}^{k} \mathbf{z}_{j_{l}} \\
        &= \frac{1}{k} \sum_{i=1}^{k} \left( \left( -\mathbf{1} \right)^{i+1} \odot \sum_{j=1}^{m} \mathbf{z}_{j}^{i} \odot \sum_{j_{1} > \ldots > j_{k-i}}^{m} \bigodot_{l=1}^{k-i} \mathbf{z}_{j_{l}} \right) \\
        &= \frac{1}{k} \sum_{i=1}^{k} \left( \left( -1 \right)^{i+1} g_{1}(\mathbf{Z}^{i}) \odot g_{k-i}(\mathbf{Z}) \right).
    \end{split}
\end{equation}\noindent
This implies that the proposed recursive form satisfies \textbf{Definition A.1}. \hfill $\square$

\section*{Acknowledgment}
This research was supported by a grant of the Korea Health Technology R\&D Project through the Korea Health Industry Development Institute (KHIDI), funded by the Ministry of Health \& Welfare, Republic of Korea (grant number: RS-2021-KH114109).

\bibliographystyle{ieeetr}
\bibliography{references}

@article{Yan2022,
    author	=	{Yan, Bencheng and Wang, Pengjie and Zhang, Kai and Li, Feng and Xu, Jian and Zheng, Bo},
    title	=	{{APG: Adaptive Parameter Generation Network for Click-Through Rate Prediction}},
    year	=	{2022},
    journal	=	{arXiv preprint arXiv:2203.16218},
}

@inproceedings{Chen2016,
    author	=	{Chen, Tianqi and Guestrin, Carlos},
    title	=	{{XGBoost: A Scalable Tree Boosting System}},
    year	=	{2016},
    booktitle	=	{Proceedings of the 22nd ACM SIGKDD International Conference on Knowledge Discovery and Data Mining},
}

@article{Dal2015,
    author	=	{Dal Pozzolo, Andrea},
    title	=	{{Adaptive Machine Learning for Credit Card Fraud Detection}},
    year	=	{2015},
    journal	=	{PhD Thesis, Universit{\'e} libre de Bruxelles},
}

@article{johnson2016mimic,
  title={{MIMIC-III, A Freely Accessible Critical Care Database}},
  author={Johnson, Alistair EW and Pollard, Tom J and Shen, Lu and Lehman, Li-wei H and Feng, Mengling and Ghassemi, Mohammad and Moody, Benjamin and Szolovits, Peter and Anthony Celi, Leo and Mark, Roger G},
  journal={Scientific data},
  volume={3},
  number={1},
  pages={1--9},
  year={2016}
}

@article{popov2019neural,
  title={{Neural Oblivious Decision Ensembles for Deep Learning on Tabular Data}},
  author={Popov, Sergei and Morozov, Stanislav and Babenko, Artem},
  journal={arXiv preprint arXiv:1909.06312},
  year={2019}
}

@online{california,
  title={{California Housing Prices}},
  howpublished="\url{https://www.kaggle.com/datasets/camnugent/california-housing-prices}",
  note="[Accessed: 05 Jun 2025]"
}

@online{insurance,
  title={{Insurance}},
  howpublished="\url{https://www.kaggle.com/datasets/mirichoi0218/insurance}",
  note="[Accessed: 05 Jun 2025]"
}

@online{house,
  title={{House Prices}},
  howpublished="\url{https://www.kaggle.com/competitions/house-prices-advanced-regression-techniques}",
  note="[Accessed: 05 Jun 2025]"
}

@online{bike,
  title={{Bikeshare}},
  howpublished="\url{https://archive.ics.uci.edu/ml/datasets/bike+sharing+dataset}",
  note="[Accessed: 05 Jun 2025]"
}

@online{year,
  title={{Year}},
  howpublished="\url{https://archive.ics.uci.edu/ml/datasets/yearpredictionmsd}",
  note="[Accessed: 05 Jun 2025]"
}

@online{fico,
  title={{FICO}},
  howpublished="\url{https://www.kaggle.com/datasets/averkiyoliabev/home-equity-line-of-creditheloc}",
  note="[Accessed: 05 Jun 2025]"
}

@online{support2,
  title={{SUPPORT2}},
  howpublished="\url{https://archive.ics.uci.edu/dataset/880/support2}",
  note="[Accessed: 05 Jun 2025]"
}

@online{click,
  title={{Click}},
  howpublished="\url{https://www.kaggle.com/c/kddcup2012-track2}",
  note="[Accessed: 05 Jun 2025]"
}

@inproceedings{gui2019afs,
    title="{AFS: An Attention-based Mechanism for Supervised Feature Selection}",
    author={Gui, Ning and Ge, Danni and Hu, Ziyin},
    booktitle={Proceedings of the AAAI Conference on Artificial Intelligence},
    year={2019},
}

@inproceedings{vskrlj2020feature,
    title="{Feature Importance Estimation with Self-Attention Networks}",
    author={{\v{S}}krlj, Bla{\v{z}} and D{\v{z}}eroski, Sa{\v{s}}o and Lavra{\v{c}}, Nada and Petkovi{\v{c}}, Matej},
    booktitle={24th European Conference on Artificial Intelligence},
    year={2020},
}

@inproceedings{arik2021tabnet,
    title="{TabNet: Attentive Interpretable Tabular Learning}",
    author={Ar{\i}k, Sercan O and Pfister, Tomas},
    booktitle={Proceedings of the AAAI Conference on Artificial Intelligence},
    year={2021},
}

@inproceedings{serrano2019attention,
    title="{Is Attention Interpretable?}",
    author={Serrano, Sofia and Smith, Noah A.},
    booktitle={Proceedings of the 57th Annual Meeting of the Association for Computational Linguistics},
    year={2019},
}

@inproceedings{grimsley2020attention,
    title="{Why Attention is Not Explanation: Surgical Intervention and Causal Reasoning about Neural Models}",
    author={Grimsley, Christopher and Mayfield, Elijah and R.S. Bursten, Julia},
    booktitle={Proceedings of the Twelfth Language Resources and Evaluation Conference},
    year={2020},
}

@inproceedings{tutek2020staying,
    title={Staying True to Your Word: (How) Can Attention Become Explanation?},
    author={Tutek, Martin and Snajder, Jan},
    booktitle={Proceedings of the 5th Workshop on Representation Learning for NLP},
    year={2020},
}

@article{bach2015pixel,
    title="{On Pixel-Wise Explanations for Non-Linear Classifier Decisions by Layer-Wise Relevance Propagation}",
    author={Bach, Sebastian and Binder, Alexander and Montavon, Gr{\'e}goire and Klauschen, Frederick and M{\"u}ller, Klaus-Robert and Samek, Wojciech},
    journal = {PLOS ONE},
    volume = {10},
    number = {7},
    pages = {1-46},
    year = {2015},
}

@inproceedings{shrikumar2017learning,
    title="{Learning Important Features Through Propagating Activation Differences}",
    author={Shrikumar, Avanti and Greenside, Peyton and Kundaje, Anshul},
    booktitle={International Conference on Machine Learning},
    year={2017},
}

@inproceedings{ribeiro2016why,
    title="{"Why Should I Trust You?" Explaining the Predictions of Any Classifier}",
    author={Marco Tulio Ribeiro and Sameer Singh and Carlos Guestrin},
    booktitle={Proceedings of the 22nd ACM SIGKDD International Conference on Knowledge Discovery and Data Mining},
    year={2016},
}

@inproceedings{lundberg2017unified,
    author={Lundberg, Scott M and Lee, Su-In},
    title="{A Unified Approach to Interpreting Model Predictions}",
    booktitle={Proceedings of the 30th International Conference on Neural Information Processing Systems},
    year={2017},
}

@article{rudin2019stop,
    title="{Stop explaining black box machine learning models for high stakes decisions and use interpretable models instead}",
    author={Rudin, Cynthia},
    journal={Nature Machine Intelligence},
    volume={1},
    number={5},
    pages={206--215},
    year={2019},
}

@inproceedings{rudin2018please,
    title="{Please Stop Explaining Black Box Models for High Stakes Decisions}",
    author={Rudin, Cynthia},
    booktitle={32nd Conference on Neural Information Processing Systems (NIPS 2018), Workshop on Critiquing and Correcting Trends in Machine Learning},
    year={2018},
}

@article{hastie1987generalized,
    title	=	"{Generalized Additive Models: Some Applications}",
    author	=	{Hastie, Trevor and Tibshirani, Robert},
    journal	=	{Journal of the American Statistical Association},
    volume	=	{82},
    number  =   {398},
    pages	=	{371--386},
    year	=	{1987},
}

@inproceedings{lou2012intelligible,
    title   	=	"{Intelligible Models for Classification and Regression}",
    author  	=	{Lou, Yin and Caruana, Rich and Gehrke, Johannes},
    booktitle	=	{Proceedings of the 18th ACM SIGKDD International Conference on Knowledge Discovery and Data Mining},
    year    	=	{2012},
}

@inproceedings{chang2021interpretable,
    title       =   "{How Interpretable and Trustworthy are GAMs?}",
    author      =   {Chang, Chun-Hao and Tan, Sarah and Lengerich, Ben and Goldenberg, Anna and Caruana, Rich},
    booktitle   =   {Proceedings of the 27th ACM SIGKDD Conference on Knowledge Discovery \& Data Mining},
    year        =   {2021},
}

@inproceedings{caruana2015intelligible,
    title       =   "{Intelligible Models for HealthCare: Predicting Pneumonia Risk and Hospital 30-day Readmission}",
    author      =   {Caruana, Rich and Lou, Yin and Gehrke, Johannes and Koch, Paul and Sturm, Marc and Elhadad, Noemie},
    booktitle   =   {Proceedings of the 21st ACM SIGKDD international conference on knowledge discovery and data mining},
    year        =   {2015},
}

@inproceedings{lou2013accurate,
    title   	=	"{Accurate Intelligible Models with Pairwise Interactions}",
    author  	=	{Lou, Yin and Caruana, Rich and Gehrke, Johannes and Hooker, Giles},
    booktitle	=	{Proceedings of the 19th ACM SIGKDD International Conference on Knowledge Discovery and Data Mining},
    year    	=	{2013},
}

@article{nori2019interpret,
    title	=	"{InterpretML: A Unified Framework for Machine Learning Interpretability}",
    author	=	{Nori, Harsha and Jenkins, Samuel and Koch, Paul and Caruana, Rich},
    journal	=	{arXiv preprint arXiv:1909.09223},
    year	=	{2019},
}

@inproceedings{potts1999generalized,
    title       =   "{Generalized Additive Neural Networks}",
    author      =   {Potts, William JE},
    booktitle   =   {Proceedings of the fifth ACM SIGKDD international conference on Knowledge discovery and data mining},
    year        =   {1999},
}

@inproceedings{agarwal2020neural,
    title   	=	"{Neural Additive Models: Interpretable Machine Learning with Neural Nets}",
    author  	=	{Rishabh Agarwal and Levi Melnick and Nicholas Frosst and Xuezhou Zhang and Ben Lengerich and Rich Caruana and Geoffrey Hinton},
    booktitle  	=	{Proceedings of the 34th International Conference on Neural Information Processing Systems},
    year        =	{2020},
}

@inproceedings{chang2022node,
    title       =   "{NODE-GAM: Neural Generalized Additive Model for Interpretable Deep Learning}",
    author      =   {Chang, Chun-Hao and Caruana, Rich and Goldenberg, Anna},
    booktitle   =   {International Conference on Learning Representations},
    year        =   {2022},
}

@inproceedings{rendle2010factorization,
    title       =	"{Factorization Machines}",
    author      =	{Rendle, Steffen},
    booktitle	=	{2010 IEEE International Conference on Data Mining},
    year        =	{2010},
}

@inproceedings{blondel2016higer,
    title   	=	"{Higher-Order Factorization Machines}",
    author  	=	{Mathieu Blondel and Akinori Fujino and Naonori Ueda and Masakazu Ishihata},
    booktitle	=	{Proceedings of the 29th International Conference on Neural Information Processing Systems},
    year    	=	{2016},
}

@inproceedings{xiao2017attentional,
    title       =   "{Attentional Factorization Machines: Learning the Weight of Feature Interactions via Attention Networks}",
    author      =   {Xiao, Jun and Ye, Hao and He, Xiangnan and Zhang, Hanwang and Wu, Fei and Chua, Tat-Seng},
    booktitle   =   {Proceedings of the 26th International Joint Conference on Artificial Intelligence},
    year        =   {2017},
}

@inproceedings{cheng2016wide,
    title   	=	"{Wide \& Deep Learning for Recommender Systems}",
    author  	=	{Cheng, Heng-Tze and Koc, Levent and Harmsen, Jeremiah and Shaked, Tal and Chandra, Tushar and Aradhye, Hrishi and Anderson, Glen and Corrado, Greg and Chai, Wei and Ispir, Mustafa and others},
    booktitle	=	{Proceedings of the 1st Workshop on Deep Learning for Recommender Systems},
    year    	=	{2016},
}

@inproceedings{guo2017deepfm,
    title   	=	"{DeepFM: A Factorization-Machine Based Neural Network for CTR Prediction}",
    author  	=	{Guo, Huifeng and Tang, Ruiming and Ye, Yunming and Li, Zhenguo and He, Xiuqiang},
    booktitle	=	{Proceedings of the 26th International Joint Conference on Artificial Intelligence},
    year    	=	{2017},
}

@inproceedings{wang2017deep,
    title       =   "{Deep \& Cross Network for Ad Click Predictions}",
    author      =   {Wang, Ruoxi and Fu, Bin and Fu, Gang and Wang, Mingliang},
    booktitle   =   {Proceedings of the ADKDD'17},
    year        =   {2017},
}

@inproceedings{lian2018xdeepfm,
    title   	=	"{xDeepFM: Combining Explicit and Implicit Feature Interactions for Recommender Systems}",
    author  	=	{Lian, Jianxun and Zhou, Xiaohuan and Zhang, Fuzheng and Chen, Zhongxia and Xie, Xing and Sun, Guangzhong},
    booktitle	=	{Proceedings of the 24th ACM SIGKDD International Conference on Knowledge Discovery \& Data Mining},
    year    	=	{2018},
}

@inproceedings{kim2020combining,
    title   	=	"{Combining Multiple Implicit-Explicit Interactions for Regression Analysis}",
    author  	=	{Kim, Minkyu and Lee, Suan and Kim, Jinho},
    booktitle	=	{2020 IEEE International Conference on Big Data (Big Data)},
    year    	=	{2020},
}

@inproceedings{cheng2020adaptive,
    title   	=	"{Adaptive Factorization Network: Learning Adaptive-Order Feature Interactions}",
    author  	=	{Cheng, Weiyu and Shen, Yanyan and Huang, Linpeng},
    booktitle	=	{Proceedings of the AAAI Conference on Artificial Intelligence},
    year    	=	{2020},
}

@inproceedings{tolomei2017interpretable,
  title={Interpretable Predictions of Tree-based Ensembles via Actionable Feature Tweaking},
  author={Tolomei, Gabriele and Silvestri, Fabrizio and Haines, Andrew and Lalmas, Mounia},
  booktitle={Proceedings of the 23rd ACM SIGKDD International Conference on Knowledge Discovery and Data Mining},
  pages={465--474},
  year={2017}
}

@inproceedings{mothilal2020explaining,
  title={Explaining Machine Learning Classifiers through Diverse Counterfactual Explanations},
  author={Mothilal, Ramaravind K and Sharma, Amit and Tan, Chenhao},
  booktitle={Proceedings of the 2020 Conference on Fairness, Accountability, and Transparency},
  pages={607--617},
  year={2020}
}

@inproceedings{karimi2020model,
  title={Model-Agnostic Counterfactual Explanations for Consequential Decisions},
  author={Karimi, Amir-Hossein and Barthe, Gilles and Balle, Borja and Valera, Isabel},
  booktitle={International Conference on Artificial Intelligence and Statistics},
  pages={895--905},
  year={2020}
}

@inproceedings{lucic2022focus,
  title={FOCUS: Flexible Optimizable Counterfactual Explanations for Tree Ensembles},
  author={Lucic, Ana and Oosterhuis, Harrie and Haned, Hinda and de Rijke, Maarten},
  booktitle={Proceedings of the AAAI Conference on Artificial Intelligence},
  volume={36},
  number={5},
  pages={5313--5322},
  year={2022}
}

@inproceedings{chen2022relax,
  title={ReLAX: Reinforcement Learning Agent Explainer for Arbitrary Ppredictive Models},
  author={Chen, Ziheng and Silvestri, Fabrizio and Wang, Jia and Zhu, He and Ahn, Hongshik and Tolomei, Gabriele},
  booktitle={Proceedings of the 31st ACM International Conference on Iinformation \& Knowledge Management},
  pages={252--261},
  year={2022}
}

@inproceedings{filip_radenovic_neural_2022,
	title = {Neural {Basis} {Models} for {Interpretability}},
	booktitle = {Advances in {Neural} {Information} {Processing} {Systems}},
	author = {{Filip Radenovic} and {Abhimanyu Dubey} and {Dhruv Mahajan}},
	year = {2022},
}

@inproceedings{xu_sparse_2023,
	title = {Sparse {Neural} {Additive} {Model}: {Interpretable} {Deep} {Learning} with {Feature} {Selection} via {Group} {Sparsity}},
	booktitle = {Joint {European} {Conference} on {Machine} {Learning} and {Knowledge} {Discovery} in {Databases}},
	author = {Xu, Shiyun and Bu, Zhiqi and Chaudhari, Pratik and Barnett, Ian J.},
	year = {2023},
}

@inproceedings{tan_distill-and-compare_2018,
	title = {Distill-and-{Compare}: {Auditing} {Black}-{Box} {Models} {Using} {Transparent} {Model} {Distillation}},
	booktitle = {Proceedings of the 2018 {AAAI}/{ACM} {Conference} on {AI}, {Ethics}, and {Society}},
	author = {Tan, Sarah and Caruana, Rich and Hooker, Giles and Lou, Yin},
	year = {2018},
}

@article{pedersen_hierarchical_2019,
	title = {Hierarchical generalized additive models in ecology: an introduction with mgcv},
	volume = {7},
	journal = {PeerJ},
	author = {Pedersen, Eric J. and Miller, David L. and Simpson, Gavin L. and Ross, Noam},
	month = may,
	year = {2019},
	pages = {e6876},
}

@article{hastie_generalized_1995,
	title = {Generalized additive models for medical research},
	volume = {4},
	number = {3},
	journal = {Statistical Methods in Medical Research},
	author = {Hastie, Trevor and Tibshirani, Robert},
	month = sep,
	year = {1995},
	pages = {187--196},
}

@phdthesis{akhavan_evaluating_2025,
	title = {Evaluating the {Faithfulness} of {Local} {Feature} {Attribution} {Explanations}: {Can} {We} {Trust} {Explainable} {AI}?},
	school = {KTH Royal Institute of Technology},
	author = {Akhavan Rahnama, Amir Hossein},
	year = {2025},
}

@inproceedings{rahnama_study_2019,
	title = {A study of data and label shift in the {LIME} framework},
	booktitle = {Workshop on {Human}-{Centric} {Machine} {Learning} at the 33rd {Conference} on {Neural} {Information} {Processing} {Systems}},
	author = {Rahnama, Amir Hossein Akhavan and Boström, Henrik},
	year = {2019},
}

@article{montavon_methods_2018,
	title = {Methods for interpreting and understanding deep neural networks},
	volume = {73},
	journal = {Digital Signal Processing},
	author = {Montavon, Grégoire and Samek, Wojciech and Müller, Klaus-Robert},
	year = {2018},
	pages = {1--15},
}

@inproceedings{ghorbani_interpretation_2019,
	title = {Interpretation of {Neural} {Networks} {Is} {Fragile}},
	booktitle = {Proceedings of the {AAAI} {Conference} on {Artificial} {Intelligence}},
	author = {Ghorbani, Amirata and Abid, Abubakar and Zou, James},
	year = {2019},
}

@article{rahnama_can_2024,
	title = {Can local explanation techniques explain linear additive models?},
	volume = {38},
	journal = {Data Mining and Knowledge Discovery},
	author = {Rahnama, Amir Hossein Akhavan and Bütepage, Judith and Geurts, Pierre and Boström, Henrik},
	year = {2024},
	pages = {237--280},
}

@inproceedings{liu_synthetic_2021,
	title = {Synthetic {Benchmarks} for {Scientiﬁc} {Research} in {Explainable} {Machine} {Learning}},
	booktitle = {Advances in {Neural} {Information} {Processing} {Systems}},
	author = {Liu, Yang and Khandagale, Sujay and White, Colin and Neiswanger, Willie},
	year = {2021},
}

@inproceedings{akhavan_rahnama_blame_2023,
    title = {The {Blame} {Problem} in {Evaluating} {Local} {Explanations} and {How} to {Tackle} {It}},
    booktitle = {European {Conference} on {Artificial} {Intelligence}},
    author = {Akhavan Rahnama, Amir Hossein},
    year = {2023},
}

\end{document}